\documentclass[3p,preprint,12pt]{elsarticle} 
\makeatletter\if@twocolumn\PassOptionsToPackage{switch}{lineno}\else\fi\makeatother

\usepackage[english]{babel}
\usepackage[T1]{fontenc}        
\usepackage[utf8]{inputenc}     

\usepackage{setspace}           
\usepackage{xspace}             

\usepackage{comment}            
\usepackage{lineno}             
\usepackage{hyperref}

\usepackage{multibib}                   
\usepackage{graphicx}

\usepackage[mathscr]{eucal}
\usepackage{amsmath}
\usepackage{amssymb}            
\usepackage{amsfonts}           

\usepackage{hyperref}
\usepackage{cleveref}           

\usepackage{tikz}               
\usepackage{tikzsymbols}        

\usepackage{xcolor}
\usepackage{xargs}
\usepackage{soul}

\usepackage{subcaption}
\usepackage{float}
\usepackage{booktabs}
\usepackage{tabularx}
\usepackage{multirow}

\usepackage{algorithm} 
\usepackage{algpseudocode} 

\newcommand{\ie}{i.\,e.\xspace}
\newcommand{\eg}{e.\,g.\xspace}

\newtheorem{theorem}{Theorem}

\newtheorem{proposition}[theorem]{Proposition}

\makeatletter
\DeclareRobustCommand{\qed}{%
  \ifmmode 
  \else \leavevmode\unskip\penalty9999 \hbox{}\nobreak\hfill
  \fi
  \quad\hbox{\qedsymbol}}
\newcommand{\openbox}{\leavevmode
  \hbox to.77778em{%
  \hfil\vrule
  \vbox to.675em{\hrule width.6em\vfil\hrule}%
  \vrule\hfil}}
\newcommand{\qedsymbol}{\openbox}
\newenvironment{proof}[1][\proofname]{\par
  \normalfont
  \topsep6\p@\@plus6\p@ \trivlist
  \item[\hskip\labelsep\itshape
    #1.]\ignorespaces
}{%
  \qed\endtrivlist
}
\newcommand{\proofname}{Proof}
\makeatother

\usepackage{tabulary}
\makeatletter
\let\save@ps@pprintTitle\ps@pprintTitle
\def\ps@pprintTitle{\save@ps@pprintTitle\gdef\@oddfoot{\footnotesize\itshape \null\hfill\today}}
\def\hlinewd#1{%
  \noalign{\ifnum0=`}\fi\hrule \@height #1%
  \futurelet\reserved@a\@xhline}

\AtBeginDocument{\ifNAT@numbers \biboptions{sort&compress}\fi}
\makeatother

  \renewenvironment{abstract}{\global\setbox\absbox=\vbox\bgroup
    \hsize=\textwidth%
  \noindent\unskip\textbf{}
   \par\medskip\noindent\unskip\ignorespaces}
   {\egroup}
  
\usepackage{ifluatex}
\ifluatex
\usepackage{fontspec}
\defaultfontfeatures{Ligatures=TeX}
\usepackage[]{unicode-math}
\unimathsetup{math-style=TeX}
\else 
\usepackage[utf8]{inputenc}
\fi 
\ifluatex\else\usepackage{stmaryrd}\fi

\usepackage{url,multirow,morefloats,floatflt,cancel,tfrupee}
\makeatletter

\AtBeginDocument{\@ifpackageloaded{textcomp}{}{\usepackage{textcomp}}}
\makeatother
\usepackage{colortbl}
\usepackage{xcolor}
\usepackage{pifont}
\usepackage[nointegrals]{wasysym}
\makeatletter

\def\mcWidth#1{\csname TY@F#1\endcsname+\tabcolsep}

\def\cAlignHack{\rightskip\@flushglue\leftskip\@flushglue\parindent\z@\parfillskip\z@skip}
\def\rAlignHack{\rightskip\z@skip\leftskip\@flushglue \parindent\z@\parfillskip\z@skip}

\@ifundefined{etal}{}{}

\usepackage{ifxetex}
\ifxetex\else\if@twocolumn\@ifpackageloaded{stfloats}{}{\usepackage{dblfloatfix}}\fi\fi

\AtBeginDocument{
\expandafter\ifx\csname eqalign\endcsname\relax
\def\eqalign#1{\null\vcenter{\def\\{\cr}\openup\jot\m@th
  \ialign{\strut$\displaystyle{##}$\hfil&$\displaystyle{{}##}$\hfil
      \crcr#1\crcr}}\,}
\fi
}

\AtBeginDocument{%
  \@ifpackageloaded{endfloat}%
   {\renewcommand\efloat@iwrite[1]{\immediate\expandafter\protected@write\csname efloat@post#1\endcsname{}}}{\newif\ifefloat@tables}%
}%

\def\BreakURLText#1{\@tfor\brk@tempa:=#1\do{\brk@tempa\hskip0pt}}
\let\lt=<
\let\gt=>
\def\processVert{\ifmmode|\else\textbar\fi}

\@ifundefined{subparagraph}{
\def\subparagraph{\@startsection{paragraph}{5}{2\parindent}{0ex plus 0.1ex minus 0.1ex}%
{0ex}{\normalfont\small\itshape}}%
}{}

\newcommand\role[1]{\unskip}
\newcommand\aucollab[1]{\unskip}
  
\@ifundefined{tsGraphicsScaleX}{\gdef\tsGraphicsScaleX{1}}{}
\@ifundefined{tsGraphicsScaleY}{\gdef\tsGraphicsScaleY{.9}}{}
\def\checkGraphicsWidth{\ifdim\Gin@nat@width>\linewidth
	\tsGraphicsScaleX\linewidth\else\Gin@nat@width\fi}

\def\checkGraphicsHeight{\ifdim\Gin@nat@height>.9\textheight
	\tsGraphicsScaleY\textheight\else\Gin@nat@height\fi}

\def\fixFloatSize#1{}
\let\ts@includegraphics\includegraphics

\def\inlinegraphic[#1]#2{{\edef\@tempa{#1}\edef\baseline@shift{\ifx\@tempa\@empty0\else#1\fi}\edef\tempZ{\the\numexpr(\numexpr(\baseline@shift*\f@size/100))}\protect\raisebox{\tempZ pt}{\ts@includegraphics{#2}}}}

\AtBeginDocument{\def\includegraphics{\@ifnextchar[{\ts@includegraphics}{\ts@includegraphics[width=\checkGraphicsWidth,height=\checkGraphicsHeight,keepaspectratio]}}}

\DeclareMathAlphabet{\mathpzc}{OT1}{pzc}{m}{it}

\def\URL#1#2{\@ifundefined{href}{#2}{\href{#1}{#2}}}

\def\UrlOrds{\do\*\do\-\do\~\do\'\do\"\do\-}%
\g@addto@macro{\UrlBreaks}{\UrlOrds}

\edef\fntEncoding{\f@encoding}

\makeatother

\newif\ifmultipleabstract\multipleabstractfalse%
\newcounter{count}

\newcommand{\CC}{C\nolinebreak\hspace{-.05em}\raisebox{.4ex}{\tiny\bf +}\nolinebreak\hspace{-.10em}\raisebox{.4ex}{\tiny\bf +}}
\def\CC{{C\nolinebreak[4]\hspace{-.05em}\raisebox{.4ex}{\tiny\bf ++}}}

\newcommand{\probFullName}{maximum customers' satisfaction problem with car-sharing demand pairs between two stations}
\newcommand{\probInitials}{car-sharing problem}

\newcommand{\nA}{m_{^{_{\hspace*{-.1em}\mathscr{A}}}}\hspace*{-.2em}}
\newcommand{\nB}{m_{^{_{\hspace*{-.1em}\mathscr{B}}}}\hspace*{-.2em}}
\newcommand{\TA}{\mathscr{T}_{\hspace*{-.2em}\mathscr{A}}}
\newcommand{\TB}{\mathscr{T}_{\hspace*{-.1em}\mathscr{B}}}

\newcommand{\A}{$\mathscr{A}$}
\newcommand{\B}{$\mathscr{B}$}
\newcommand{\W}{$\mathscr{W}$}
\newcommand{\Z}{$\mathscr{Z}$}
\newcommand{\UB}{U\!B}
\newcommand{\LB}{L\!B}

\newcommand{\outbound}[1]{{o}_{#1}}
\newcommand{\return}[1]{{r}_{#1}}

\newcommand{\nMax}[1]{{#1}_{max}}

\addto\captionsenglish{}

\journal{Computers \& Operations Research}

\begin{document}

    \begin{frontmatter}

        \title{Maximum Customers' Satisfaction in One-way Car-sharing:\\Modeling, Exact and Heuristic Solving}
        \author{Welverton R. Silva}
        \ead{welverton.silva@ic.unicamp.br}
        \author{Rafael C.\,S. Schouery}
        \ead{rafael@ic.unicamp.br}
        \address{Institute of Computing, University of Campinas\\Campinas, São Paulo, Brazil}
         
        \begin{abstract}
            \noindent One-way car-sharing systems are transportation systems that allow customers to rent cars at stations scattered around the city, use them for a short journey, and return them at any station.
            The maximum customers' satisfaction problem concerns the task of assigning the cars, initially located at given stations, to maximize the number of satisfied customers. We consider the problem with two stations where each customer has exactly two demands in opposite directions between both stations, and a customer is satisfied only if both their demands are fulfilled.
            For solving this problem, we propose mixed-integer programming~(MIP) models and matheuristics based on local search. We created a benchmark of instances used to test the exact and heuristic approaches. Additionally, we proposed a preprocessing procedure to reduce the size of the instance. 
            Our MIP models can solve to optimality~$85\%$ of the proposed instances with~$1000$ customers in~$10$ minutes, with an average gap smaller than~$0.1\%$ for all these instances. For larger instances ($2500$ and $5000$ customers), except for some particular cases, they presented an average gap smaller than~$0.8\%$. Also, our local-based matheuristics presented small average gaps which are better than the MIP models in some larger instances.
        \end{abstract}
        
    
    \end{frontmatter}
    
    \section{Introduction}
\label{sec:introduction}

Car-sharing is a service that provides car rentals where customers occasionally rent cars for short periods. The most common car-sharing services are station-based systems, which designate stations in certain zones of the city. A station-based system is a \textit{two-way} or \textit{one-way} car-sharing system. In a two-way car-sharing system, the customers must return the car at the same departure station, while in a one-way car-sharing system the customers can return the car at a chosen station, as long as the drop-off station and time is indicated in advance~\cite{Nourinejad2015}. In this paper, we focus on one-way station-based car-sharing systems, in particular systems where all driving demands are known in advance.

The fact that a customer does not return the car to the origin station can generate an imbalance across the origin and destination stations. Thus, most researches on one-way car-sharing have studies the balancing problem of the car fleet, also called the relocation problem  (\eg, Ait-Ouahmed et al.~\cite{Ait-Ouahmed2018}; Boyac{\i} et al.~\cite{Boyaci2015, Boyaci2017}; Gambella et al.~\cite{Gambella2018}; Jorge~et~al.~\cite{Jorge2012}; Nourinejad et al.~\cite{Nourinejad2015}; and Zhao et al.~\cite{Zhao2018}). Also, other combinatorial optimization problems are considered, such as the problems of defining the number of car-sharing stations, location, and fleet size (\eg George and Xia~\cite{George2011}).

In this paper, we consider a combinatorial optimization problem, proposed by B{\"o}hmov{\'a} et al.~\cite{Bohmova2016}, that explores the main idea in the one-way car-sharing, \ie, the flexible drop-off idea in the case where the driving demands are known in advance. We consider the problem with two stations where each customer has exactly two demands in opposite directions between both stations. In this problem, the trajectory of the car in the rental period is irrelevant, and it can be abstracted as moving the car from one station to another during a given time interval. The problem aims to determine the maximum number of customers that can be satisfied by the existing fleet of cars distributed within these two stations. We say that a customer is \textit{satisfied} if and only if both the demands are fulfilled.

This problem and some variants are already considered in the literature, but this is the first work to consider the design of exact algorithms and heuristics for this problem. 

From an approximation algorithm standpoint, B{\"o}hmov{\'a} et al.~\cite{Bohmova2016} show that maximizing the number of satisfied customers is NP-hard and, in fact, APX-hard, even for the particular case where there is only a single car available and all demands have the same duration. Later on, 
Luo et al.~\cite{Luo2018a, Luo2018b, Luo2018c} and Luo et al.~\cite{Luo2019} also considered the problem of maximizing customers' satisfaction in one-way car-sharing systems, presenting and analyzing online algorithms for the same problem that we consider and some close related variants (for example, when the customers have only one demand). 

A practical application for the problem considered in this paper arises from one-way car-sharing systems like the Intelligent Sharing System (SCI, \textit{Sistema de Compartilhamento Inteligente}) of electric cars, an initiative of Itaipu Binacional (a hydroelectric power plant owned by Brazil and Paraguay) and Itaipu Technological Park, in partnership with the Center for Engineering and Innovation of Mobility Industries (CEiiA). The SCI serves the employees in transportation within the Brazilian margin of the plant, which includes electric car pick-up and drop-off stations in different internal buildings of the company~\cite{SCI2017}.

\medskip

The remainder of the paper is organized as follows. Section~\ref{sec:problem} describes the problem formally, and how instances of the problem can be seen as network flow instances. Section~\ref{sec:preprocessing} presents the proposed preprocessing procedure to reduce the instance's size. Sections~\ref{sec:mip} and~\ref{sec:heuristics} present an exact method by using a mixed-integer programming model and the proposed matheuristic algorithms to solve our car-sharing problem, respectively. Section~\ref{sec:results} reports and discusses the experimental computational results, and Section~\ref{sec:conclusion} presents our final conclusions.

    \section{Problem statement}
\label{sec:problem}

As previously mentioned, B{\"o}hmov{\'a} et al.~\cite{Bohmova2016} propose a problem motivated in one-way car-sharing. In the same way that the problem was described by these authors, we formally define the problem with only two stations~{\A} and~{\B}, with an initial distribution of indistinguishable cars within these stations. There are~$n$ customers and each of them has exactly two demands, one from station~{\A} to station~{\B} and the other from station~{\B} to station~{\A}, but not necessarily in this order. The \emph{\probFullName}, which we will simply refer to it as the \emph{\probInitials}, can be formally defined as follows.

Consider $\TA\!=\!\{a_1,\dots,\nMax{a}\}$ and $\TB\!=\!\{b_1,\dots,\nMax{b}\}$ as finite sets of discrete time instants for  \emph{stations} {\A} and {\B}, respectively, such that $\TA \cup \TB$ is totally-ordered. Let $\mathcal{C}\!=\!\{(\outbound{c}, \return{c}): 1 \leq c \leq n \}$ be a set of \emph{customers}, where each customer is an ordered pair of \emph{demands} such that either $\outbound{c}\!=\!(a_i, b_j) \in \TA\times\TB$ and $\return{c}\!=\!(b_k, a_l) \in \TB\times\TA$, with $a_i \leq b_j \leq b_k \leq a_l$, or $\outbound{c}\!=\!(b_i, a_j) \in \TB\times\TA$ and $\return{c}\!=\!(a_k, b_l) \in \TA\times\TB$, with $b_i \leq a_j \leq a_k \leq b_l$, where $\outbound{c}$ represents a request for \emph{outbound} and $\return{c}$ a request for \emph{return}. Thus, either a customer has an outbound request from \A{} to \B{} and a return request from \B{} to \A{}, or the customer has an outbound request from \B{} to \A{} and a return request from \A{} to \B{}. Let $\nA$ and $\nB$ denote the starting fleet size (total number of cars) at the stations {\A} and {\B}, respectively.

A set~$\mathcal{C}^\prime \subseteq \mathcal{C}$ is \emph{feasible} if and only if all customers can be satisfied simultaneously. A customer is \emph{satisfied} only if both their demands are fulfilled, and a demand is \emph{fulfilled} by moving one car between the stations. However, this is only possible when there is an available car at the origin station at the start time as defined by the customer's outbound request --- the rented car will be available to other customers at the destination station at the end time. The objective is to find a maximum cardinality feasible set~$\mathcal{C}^* \subseteq \mathcal{C}$.

\subsection{Network design}

In this section, we describe how instances can be modeled as capacitated directed acyclic multigraphs with two special vertices, a source~$s$ and a sink~$t$. We are interested in describing instances of the {\probInitials} as instances of a network flow problem in the same way that was described by B{\"o}hmov{\'a} et al.~\cite{Bohmova2016}.

Let $G\!=\!(V, E)$ be a network, where $V\!=\TA \cup \,\TB \cup \{s, t\}$ is the set of vertices, and~$E$ is the set of arcs. The network contains arcs of three types. All demands $\outbound{c}$ and $\return{c}$ \mbox{define} \emph{demand~arcs}. For every two consecutive vertices $a_i$, $a_{i+1}$ in $\TA$, there is a \emph{connecting arc}~$(a_i, a_{i+1})$. Similarly, there is a connecting arc for every two consecutive vertices $b_i$,~$b_{i+1}$ in~$\TB$. There are also two the connecting arcs $(\nMax{a}, t)$ and $(\nMax{b}, t)$, from the last vertices of each station to the sink. Finally, there are two \emph{source arcs} $(s, a_1)$ and $(s, b_1)$. 

To complete the construction of the network, it is necessary to assign capacity to the arcs. For all demand arcs, the capacity is set to $1$, so it follows that at most one car can be moved through the arc, and for all connecting arcs, the capacity is set to $\infty$ (or, simply,~$\nA + \nB$). Finally, the source arc $(s, a_1)$ has capacity $\nA$, and the source arc $(s, b_1)$ has capacity $\nB$. A feasible solution in this network is an integral $s$-$t$ flow that satisfies three constraints: capacity, flow conservation, and satisfying customer (\ie, whenever there is a unit flow on a demand arc of a costumer, then both demand arcs has a unit flow).

\begin{figure}[!h]

    \LARGE
    \centering
             
    \definecolor{c4}{rgb}{0.03, 0.27, 0.49}
    \definecolor{c2}{rgb}{0.81, 0.09, 0.13}
    \definecolor{c3}{rgb}{0.69, 0.40, 0.00}
    \definecolor{c1}{rgb}{0.00, 0.50, 0.00}

    \begin{subfigure}[b]{0.45\textwidth}
    \centering
    \begin{tikzpicture}[scale=.4, transform shape]
         
        \foreach \i in {0.25, 2.75} {
            \draw[ultra thick, >=stealth, ->] (0, \i) -- (16.0, \i);
        }
        
        \node[inner sep=0pt, minimum size=1.4cm] (TA) at (15.8, 3.2) {$\TA$};
        \node[inner sep=0pt, minimum size=1.2cm] (TB) at (15.8, 0.7) {$\TB$};
        
        \setcounter{count}{1}
        \foreach \i in {1, 4, 5.5, 8.5, 11.5, 14.5} {
            \def\k{\the\value{count}}
            \node[fill=gray!10, draw=black, semithick, circle, inner sep=0pt, minimum size=0.85cm] (a\k) at (\i, 2.75) {$a_{\k}$};
            \stepcounter{count}
        }
        
        \setcounter{count}{1}
        \foreach \i in {2.5, 4, 7, 8.5, 10, 13} {
            \def\k{\the\value{count}}
            \node[fill=gray!10, draw=black, semithick, circle, inner sep=0pt, minimum size=0.85cm] (b\k) at (\i, 0.25) {$b_{\k}$};
            \stepcounter{count} 
        }
    
        \draw [c3, transform canvas={xshift=0.25ex,yshift=0.15ex}, line width=0.3mm, >=stealth, ->] 
        (a1) -- (b1) node[midway, above]{\textcolor{black}{~\,$o_{3}$}};
        \draw [c3, line width=0.3mm, >=stealth, ->] 
        (b4) -- (a5) node[near start, above]{\textcolor{black}{$r_{3}$\,}};
        \draw [c2, transform canvas={xshift=-.3ex,yshift=-.15ex}, line width=0.3mm, >=stealth, ->] 
        (a1) -- (b1) node[near start, below]{\textcolor{black}{$o_{2}$\,~}};
        \draw [c2, line width=0.3mm, >=stealth, ->] 
        (b6) -- (a6) node[near start, xshift=-.08cm, yshift=0cm, above]{\textcolor{black}{$r_{2}$~}};
        \draw [c4, line width=0.3mm, >=stealth, ->] 
        (a2) -- (b3) node[midway, xshift=0.05cm, yshift=-0.1cm, above]{\textcolor{black}{\,$o_{4}$}};
        \draw [c4, line width=0.3mm, >=stealth, ->] 
        (b5) -- (a5) node[midway, xshift=-0.18cm, yshift=-0.2cm, below]{\textcolor{black}{~~\,$r_{4}$}};
        \draw [c1, line width=0.3mm, >=stealth, ->] 
        (b2) -- (a3) node[near start, xshift=-0.15cm, yshift=-0.2cm, above]{\textcolor{black}{$o_{1}$\,~}};
        \draw [c1, line width=0.3mm, >=stealth, ->] 
        (a4) -- (b6) node[near end, xshift=-0.10cm, yshift=0cm, above]{\textcolor{black}{$r_{1}$}};
    
    \end{tikzpicture}
    \end{subfigure}
    \begin{subfigure}[b]{0.45\textwidth}
    \centering
    \begin{tikzpicture}[scale=.4, transform shape]
    
        \node[fill=gray!10, draw=black, semithick, circle, inner sep=0pt, minimum size=0.85cm] (s) at (-.5,1.5) {$s$};
        \node[fill=gray!10, draw=black, semithick, circle, inner sep=0pt, minimum size=0.85cm] (t) at (16,1.5) {$t$};
        
        \setcounter{count}{1}
        \foreach \i in {1, 4, 5.5, 8.5, 11.5, 14.5} {
            \def\k{\the\value{count}}
            \node[fill=gray!10, draw=black, semithick, circle, inner sep=0pt, minimum size=0.85cm] (a\k) at (\i, 2.75) {$a_{\k}$};
            \stepcounter{count}
        }
        
        \setcounter{count}{1}
        \foreach \i in {2.5, 4, 7, 8.5, 10, 13} {
            \def\k{\the\value{count}}
            \node[fill=gray!10, draw=black, semithick, circle, inner sep=0pt, minimum size=0.85cm] (b\k) at (\i, 0.25) {$b_{\k}$};
            \stepcounter{count} 
        }
    
        \foreach \i [remember=\i as \lasti (initially 1)]  in {2, ..., 6} {
            \draw [style=dashed, line width=0.275mm, >=stealth, ->] (a\lasti) -- (a\i);
            \draw [style=dashed, line width=0.275mm, >=stealth, ->] (b\lasti) -- (b\i);
        }
        
        \draw [style=dotted, line width=0.4mm, >=stealth, ->] (s) -- (a1); 
        \draw [style=dotted, line width=0.4mm, >=stealth, ->] (s) -- (b1);
        \draw [style=dashed, line width=0.275mm, >=stealth, ->] (a6) -- (t); 
        \draw [style=dashed, line width=0.275mm, >=stealth, ->] (b6) -- (t);
     
        \draw [c3, transform canvas={xshift=0.25ex,yshift=0.15ex}, line width=0.3mm, >=stealth, ->] 
        (a1) -- (b1) node[midway, above]{\textcolor{black}{~\,$o_{3}$}};
        \draw [c3, line width=0.3mm, >=stealth, ->] 
        (b4) -- (a5) node[near start, above]{\textcolor{black}{$r_{3}$\,}};
        \draw [c2, transform canvas={xshift=-.3ex,yshift=-.15ex}, line width=0.3mm, >=stealth, ->] 
        (a1) -- (b1) node[near start, below]{\textcolor{black}{$o_{2}$\,~}};
        \draw [c2, line width=0.3mm, >=stealth, ->] 
        (b6) -- (a6) node[near start, xshift=-.08cm, yshift=0cm, above]{\textcolor{black}{$r_{2}$~}};
        \draw [c4, line width=0.3mm, >=stealth, ->] 
        (a2) -- (b3) node[midway, xshift=0.05cm, yshift=-0.1cm, above]{\textcolor{black}{\,$o_{4}$}};
        \draw [c4, line width=0.3mm, >=stealth, ->] 
        (b5) -- (a5) node[midway, xshift=-0.18cm, yshift=-0.2cm, below]{\textcolor{black}{~~\,$r_{4}$}};
        \draw [c1, line width=0.3mm, >=stealth, ->] 
        (b2) -- (a3) node[near start, xshift=-0.15cm, yshift=-0.2cm, above]{\textcolor{black}{$o_{1}$\,~}};
        \draw [c1, line width=0.3mm, >=stealth, ->] 
        (a4) -- (b6) node[near end, xshift=-0.10cm, yshift=0cm, above]{\textcolor{black}{$r_{1}$}};
        
    \end{tikzpicture}
    \end{subfigure}
    
    \caption{An example of  instance and the corresponding network flow representation. The horizontal top and bottom lines represent respectively the time instants at stations {\A} and {\B}. All driving demands are represented by arcs, where arcs of the same color indicate the demands of the same customer. Also, each label on the demand arcs specifies the demand's type.}
    \label{fig:instance_into_network}
\end{figure}
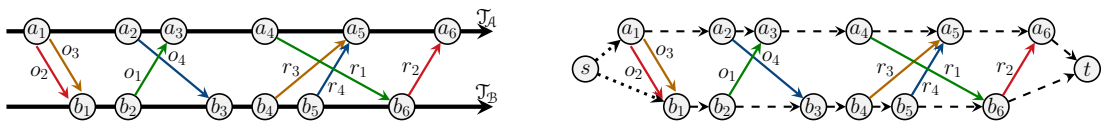

Fig.~\ref{fig:instance_into_network} illustrates a network from an arbitrary instance, where the demand arcs are displayed as full arcs, the connecting arcs are displayed as dashed arcs, and those source arcs displayed as dotted~arcs.
    \section{Preprocessing}
\label{sec:preprocessing}

Our preprocessing procedure is an iterative transformation over a given network, which is employed to improve the efficiency of our exact and heuristic algorithms. Basically, any instance modeled as a network can be transformed into another equivalent network with, potentially, a reduced number of vertices and arcs. We say that two networks are \emph{equivalent} if and only if they have the same set of feasible solutions.  

Consider a given network $G\!=\!(V,E)$ constructed as described in the previous section. We can iteratively construct a minimal network $G^\prime$ equivalent to $G$. We say that $G^\prime$ is \emph{minimal} if and only if is not possible to apply any transformation operations: (i) arc contraction, (ii) arc merge, and (iii) vertex removal.

\begin{figure}[!h]

    \LARGE
    \centering
    
    \begin{subfigure}[b]{0.45\textwidth}
    \centering
    \begin{tikzpicture}[scale=.4, transform shape]

        \definecolor{c4}{rgb}{0.03, 0.27, 0.49}
        \definecolor{c2}{rgb}{0.81, 0.09, 0.13}
        \definecolor{c3}{rgb}{0.69, 0.40, 0.00}
        \definecolor{c1}{rgb}{0.00, 0.50, 0.00}
    
        \node[fill=gray!10, draw=black, semithick, circle, inner sep=0pt, minimum size=0.85cm] (s) at (-.5,1.5) {$s$};
        \node[fill=gray!10, draw=black, semithick, circle, inner sep=0pt, minimum size=0.85cm] (t) at (16,1.5) {$t$};
        
        \node[fill=gray!10, draw=black, semithick, circle, inner sep=0pt, minimum size=0.85cm, line width=0.75pt] (a1) at (1.0, 2.75) {$a_{1}$};
        \node[fill=gray!10, draw=black, semithick, circle, inner sep=0pt, minimum size=0.85cm, line width=0.75pt] (a2) at (7.0, 2.75) {$a_{2}$};
        \node[fill=gray!10, draw=black, semithick, circle, inner sep=0pt, minimum size=0.85cm, line width=0.75pt] (a3) at (14.5, 2.75) {$a_{3}$};
        
        \node[fill=gray!10, draw=black, semithick, circle, inner sep=0pt, minimum size=0.85cm, line width=0.75pt] (b1) at (2.5, 0.25) {$b_{1}$};
        \node[fill=gray!10, draw=black, semithick, circle, inner sep=0pt, minimum size=0.85cm, line width=0.75pt] (b2) at (8.5, 0.25) {$b_{2}$};
        \node[fill=gray!10, draw=black, semithick, circle, inner sep=0pt, minimum size=0.85cm] (b3) at (13, 0.25) {$b_{3}$};
    
        \foreach \i [remember=\i as \lasti (initially 1)]  in {2, ..., 3} {
            \draw [style=dashed, line width=0.275mm, >=stealth, ->] (a\lasti) -- (a\i);
            \draw [style=dashed, line width=0.275mm, >=stealth, ->] (b\lasti) -- (b\i);
        }
        
        \draw [style=dotted, line width=0.4mm, >=stealth, ->] (s) -- (a1); 
        \draw [style=dotted, line width=0.4mm, >=stealth, ->] (s) -- (b1);
        \draw [style=dashed, line width=0.275mm, >=stealth, ->] (a3) -- (t); 
        \draw [style=dashed, line width=0.275mm, >=stealth, ->] (b3) -- (t);

        \draw [c3, transform canvas={xshift=0.25ex,yshift=0.15ex}, line width=0.3mm, >=stealth, ->] 
        (a1) -- (b1) node[midway, above]{\textcolor{black}{~\,$o_{3}$}};
        \draw [c3, transform canvas={xshift=0.25ex,yshift=-0.25ex}, line width=0.3mm, >=stealth, ->] 
        (b2) -- (a3) node[near start, xshift=0.08cm, yshift=0.05cm, below]{\textcolor{black}{$r_{3}$\,}};
        \draw [c2, transform canvas={xshift=-.3ex,yshift=-.15ex}, line width=0.3mm, >=stealth, ->] 
        (a1) -- (b1) node[near start, below]{\textcolor{black}{$o_{2}$\,~}};
        \draw [c2, line width=0.3mm, >=stealth, ->] 
        (b3) -- (a3) node[near start, xshift=-.08cm, yshift=0cm, above]{\textcolor{black}{$r_{2}$~}};
        \draw [c4, transform canvas={xshift=0.06ex, yshift=0.1ex}, line width=0.3mm, >=stealth, ->] 
        (a1) -- (b2) node[near start, xshift=0.5cm, yshift=-0.3cm, above]{\textcolor{black}{\,$o_{4}$}};
        \draw [c4, transform canvas={xshift=-0.1ex,yshift=0.2ex}, line width=0.3mm, >=stealth, ->] 
        (b2) -- (a3) node[midway, above]{\textcolor{black}{$r_{4}$}};
        \draw [c1, line width=0.3mm, >=stealth, ->] 
        (b1) -- (a2) node[near start, xshift=-0.15cm, yshift=-0.2cm, above]{\textcolor{black}{$o_{1}$~~}};
        \draw [c1, line width=0.3mm, >=stealth, ->] 
        (a2) -- (b3) node[midway, xshift=-0.10cm, yshift=0cm, above]{\textcolor{black}{$r_{1}$}};
        
    \end{tikzpicture}
    \end{subfigure}
    \caption{An example of the application of a sequence of arc contracting operations over the previous network (illustrated in Fig.~\ref{fig:instance_into_network}).
    Here, the vertices have been relabeled to simplify the presentation.}
    \label{fig:arc_contraction}
\end{figure}
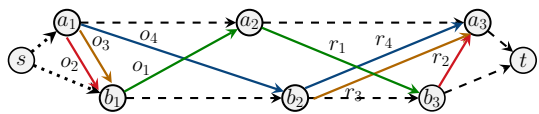

We call \emph{arc contraction} the operation that contracts  the two endpoints of a connecting arc $(x,y)$. That is, we remove vertices $x$ and $y$ (and its incident arcs), we create a new vertex $z$ representing this contraction and, except for $(x,y)$, for every removed arc $(x, w)$ (resp. $(w, y)$) we add the arc $(z, w)$ (resp. $(w, z)$) with the corresponding capacity. A connecting arc $(x,y)$ is contracted if and only if the out-degree of $x$ or the in-degree of $y$ is equal to $1$. Fig.~\ref{fig:arc_contraction} illustrates the network obtained from the previous network example (see again Fig.~\ref{fig:instance_into_network}) after some arc contractions.

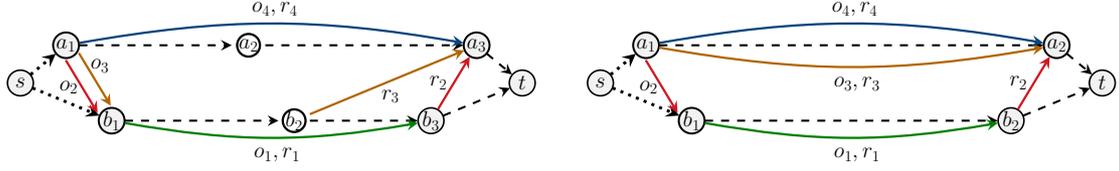
\begin{figure}[!h]

    \LARGE
    \centering
    
    \definecolor{c4}{rgb}{0.03, 0.27, 0.49}
    \definecolor{c2}{rgb}{0.81, 0.09, 0.13}
    \definecolor{c3}{rgb}{0.69, 0.40, 0.00}
    \definecolor{c1}{rgb}{0.00, 0.50, 0.00}
        
    \begin{subfigure}[b]{0.45\textwidth}
    \centering
    \begin{tikzpicture}[scale=.4, transform shape]

        \node[fill=gray!10, draw=black, semithick, circle, inner sep=0pt, minimum size=0.85cm] (s) at (-.5,1.5) {$s$};
        \node[fill=gray!10, draw=black, semithick, circle, inner sep=0pt, minimum size=0.85cm] (t) at (16,1.5) {$t$};
        
        \node[fill=gray!10, draw=black, semithick, circle, inner sep=0pt, minimum size=0.85cm, line width=0.75pt] (a1) at (1.0, 2.75) {$a_{1}$};
        \node[] (a2) at (7.0, 2.75) {$a_{2}$};
        \node[fill=gray!10, draw=black, semithick, circle, inner sep=0pt, minimum size=0.85cm, line width=0.75pt] (a3) at (14.5, 2.75) {$a_{3}$};
        
        \node[fill=gray!10, draw=black, semithick, circle, inner sep=0pt, minimum size=0.85cm, line width=0.75pt] (b1) at (2.5, 0.25) {$b_{1}$};
        \node[] (b2) at (8.5, 0.25) {$b_{2}$};
        \node[fill=gray!10, draw=black, semithick, circle, inner sep=0pt, minimum size=0.85cm] (b3) at (13, 0.25) {$b_{3}$};
    
        \foreach \i [remember=\i as \lasti (initially 1)]  in {2, ..., 3} {
            \draw [style=dashed, line width=0.275mm, >=stealth, ->] (a\lasti) -- (a\i);
            \draw [style=dashed, line width=0.275mm, >=stealth, ->] (b\lasti) -- (b\i);
        }
        
        \draw [style=dotted, line width=0.4mm, >=stealth, ->] (s) -- (a1); 
        \draw [style=dotted, line width=0.4mm, >=stealth, ->] (s) -- (b1);
        \draw [style=dashed, line width=0.275mm, >=stealth, ->] (a3) -- (t); 
        \draw [style=dashed, line width=0.275mm, >=stealth, ->] (b3) -- (t);
     
        \draw [c3, transform canvas={xshift=0.25ex,yshift=0.15ex}, line width=0.3mm, >=stealth, ->] 
        (a1) -- (b1) node[midway, above]{\textcolor{black}{~\,$o_{3}$}};
        \draw [c3, line width=0.3mm, >=stealth, ->] 
        (b2) -- (a3) node[midway, below]{\textcolor{black}{~$r_{\textnormal{3}}$}};
        \draw [c2, transform canvas={xshift=-.3ex,yshift=-.15ex}, line width=0.3mm, >=stealth, ->] 
        (a1) -- (b1) node[near start, below]{\textcolor{black}{$o_{2}$\,~}};
        \draw [c2, line width=0.3mm, >=stealth, ->] 
        (b3) -- (a3) node[near start, xshift=-.08cm, yshift=0cm, above]{\textcolor{black}{$r_{\textnormal{2}}$\,~}};
        \draw [c4, line width=0.3mm, >=stealth, ->] 
        (a1) to [bend left=10] node[midway, above]{\textcolor{black}{$~o_{\textnormal{4}},r_{\textnormal{4}}$}} (a3);
        \draw [c1, line width=0.3mm, >=stealth, ->] 
        (b1) to [bend right=10] node[midway, xshift=-0.1cm, yshift=-0.1cm, below]{\textcolor{black}{$\,~~o_{1},r_{1}$}} (b3);  
        
        \node[draw=black, semithick, circle, inner sep=0pt, minimum size=0.75cm, line width=0.85pt] (a2) at (7.0, 2.75) {};
        \node[draw=black, semithick, circle, inner sep=0pt, minimum size=0.75cm, line width=0.85pt] (b2) at (8.5, 0.25) {};
        
    \end{tikzpicture}
    \end{subfigure}
    \begin{subfigure}[b]{0.45\textwidth}
    \centering
    \begin{tikzpicture}[scale=.4, transform shape]
    
        \node[fill=gray!10, draw=black, semithick, circle, inner sep=0pt, minimum size=0.85cm] (s) at (-.5,1.5) {$s$};
        \node[fill=gray!10, draw=black, semithick, circle, inner sep=0pt, minimum size=0.85cm] (t) at (16,1.5) {$t$};
        
        \node[fill=gray!10, draw=black, semithick, circle, inner sep=0pt, minimum size=0.85cm, line width=0.75pt] (a1) at (1.0, 2.75) {$a_{1}$};
        \node[fill=gray!10, draw=black, semithick, circle, inner sep=0pt, minimum size=0.85cm, line width=0.75pt] (a2) at (14.5, 2.75) {$a_{2}$};
        
        \node[fill=gray!10, draw=black, semithick, circle, inner sep=0pt, minimum size=0.85cm, line width=0.75pt] (b1) at (2.5, 0.25) {$b_{1}$};
        \node[fill=gray!10, draw=black, semithick, circle, inner sep=0pt, minimum size=0.85cm] (b2) at (13, 0.25) {$b_{2}$};
    
        \foreach \i [remember=\i as \lasti (initially 1)]  in {2} {
            \draw [style=dashed, line width=0.275mm, >=stealth, ->] (a\lasti) -- (a\i);
            \draw [style=dashed, line width=0.275mm, >=stealth, ->] (b\lasti) -- (b\i);
        }
        
        \draw [style=dotted, line width=0.4mm, >=stealth, ->] (s) -- (a1); 
        \draw [style=dotted, line width=0.4mm, >=stealth, ->] (s) -- (b1);
        \draw [style=dashed, line width=0.275mm, >=stealth, ->] (a2) -- (t); 
        \draw [style=dashed, line width=0.275mm, >=stealth, ->] (b2) -- (t);
     
        \draw [c3, line width=0.3mm, >=stealth, ->] 
        (a1)  to [bend right=10] node[midway, xshift=-0.1cm, yshift=-0.1cm, below]{\textcolor{black}{$\,~~o_{3},r_{3}$}} (a2);
        \draw [c2, transform canvas={xshift=-.3ex,yshift=-.15ex}, line width=0.3mm, >=stealth, ->] 
        (a1) -- (b1) node[near start, below]{\textcolor{black}{$o_{2}$\,~}};
        \draw [c2, line width=0.3mm, >=stealth, ->] 
        (b2) -- (a2) node[near start, xshift=-.08cm, yshift=0cm, above]{\textcolor{black}{$r_{2}$\,~}};
        \draw [c4, line width=0.3mm, >=stealth, ->] 
        (a1) to [bend left=10] node[midway, above]{\textcolor{black}{$~o_{\textnormal{4}},r_{\textnormal{4}}$}} (a2);
        \draw [c1, line width=0.3mm, >=stealth, ->] 
        (b1) to [bend right=10] node[midway, xshift=-0.1cm, yshift=-0.1cm, below]{\textcolor{black}{$\,~~o_{1},r_{1}$}} (b2); 
        
    \end{tikzpicture}
    \end{subfigure}
    \caption{On the left, the next network obtained by a sequence of arc merging operations, and, on the right, the minimal network for the original network.} 
    \label{fig:minimal_network}
    
\end{figure}

We call \emph{arc merge} the operation that merges adjacent demand arcs of a customer. In other words, this operation merges the arcs $\outbound{c}$ and $\return{c}$ into a new arc if and only if the end-point of the outbound arc is equal to the start-point of the return arc. Note that the customer can, w.lo.g., always be satisfied with the same car in both demands. Finally, we call \emph{vertex removal} the operation that removes an expendable vertex and merges its incident arcs. A vertex is \emph{expendable} only if both its out-degree and in-degree are equal to $1$ (\eg, the vertex $a_2$ in the Fig.~\ref{fig:minimal_network} on the left side).

    \section{Mixed-integer programming}
\label{sec:mip}

In this section, we describe a mixed-integer programming~(MIP) model based on network flow assignment formulation for solving the {\probInitials}. We define a decision variable $x_e$ for each arc $e$ of a given network, where each variable indicates the amount of flow through the corresponding arc.  

The problem can be formulated as a MIP, which we denote by (CS1), as follows.
\begin{alignat}{5}
    \mathrm{(CS1)}\qquad \text{maximize}     & ~\,\sum_{c\,=1}^n x_{\outbound{c}} & \label{eq1}\\
    \text{subject to}   & \hspace{-.2em}\sum_{e\,\in\,\delta^{^{_+}}\hspace{-.2em}(v)} \hspace{-.2cm} x_e 
                             ~= \hspace{-.2em} \sum_{e\,\in\,\delta^{^{_-}}\hspace{-.2em}(v)} \hspace{-.2cm} x_e 
                                               && ~~~v \in V\!\setminus\!\!\{s,t\} \label{eq2}\\
                        & x_e \leq \nA           && ~~~e = (s, a_1\hspace{-.1em}) \label{eq3}\\
                        & x_e \leq \nB           && ~~~e = (s, b_1\hspace{-.1em}) \label{eq4}\\
                        & x_{\outbound{c}} = x_{\return{c}}  && ~~~c \in \{1,\dots, n\} \label{eq5}\\ 
                        & x_e \in \{0,1\}        && ~~~e \in \{\outbound{1},\dots, \outbound{n}, \return{1},\dots, \return{n}\} \label{eq6} \\
                        & x_e \in \mathbb{R}_{+} && ~~~e \in E\!\setminus\!\!\{\outbound{1},\dots, \outbound{n}, \return{1},\dots, \return{n} \notag \}
\end{alignat}

The objective function \eqref{eq1} maximizes the number of satisfied outbound demands, and thus, the number of satisfied customers. Constraint set \eqref{eq2} is the usual flow conservation constraints, and constraint sets \eqref{eq3}, \eqref{eq4} and \eqref{eq6} are the capacity constraints. Finally, constraint set \eqref{eq5} ensures that $\outbound{c}$ is satisfied if and only is $\return{c}$ is satisfied.

As every variable $x_{\outbound{c}}$ must always be equal to $x_{\return{c}}$, we can replace them in the model by a single decision variable $x_{d_c}$. In this way, it is possible to halve the number of binary variables. Also, the preprocessing procedure can reduce the number of real variables associated with the connecting arcs and the number of flow conservation constraints.

\subsection{Additional Constraints}

We propose a family of constraints to strengthen the formulation. These constraints can eliminate integer solutions, but, as we prove, it does not eliminate all optimal solutions. As the number of obtained inequalities can be very large, we use transitive reduction~\cite{Aho1972} to achieve a minimum equivalent set of inequalities. For this, we have defined constraints regarding priority requests, obtained from the following proposition.

\begin{proposition} Consider two customers $c\!=\!((w_i,z_j),(z_k,w_l))$ and $c^\prime\!=\!((w_{i^\prime},z_{j^\prime}),(z_{k^\prime},w_{l^\prime}))$, over any two stations {\W} and {\Z}, such that $w_i \leq w_{i^\prime}$, $z_{j^\prime} \leq z_j$, $z_k \leq z_{k^\prime}$ e $w_{l^\prime} \leq w_l$. Given a feasible set $\mathcal{C}_{1} \subseteq \mathcal{C}$ where $c \in \mathcal{C}_{1}$, and $c^\prime \notin \mathcal{C}_{1}$, the set $\mathcal{C}_{2} = \mathcal{C}_{1} \setminus \{c\} \cup \{c^\prime\}$ is also feasible.
\label{prop:inequality}
\end{proposition}

\begin{figure}[h!]

    \LARGE
    \centering
    
    \begin{tikzpicture}[scale=.4, transform shape]
        
        \centering

        \foreach \i in {0.25, 2.75} {
            \draw[ultra thick, >=stealth, ->] (0, \i) -- (15.5, \i);
        }

        \node[inner sep=0pt, minimum size=1.4cm] (TA) at (15.3, 3.2) {$\mathscr{T}_{\mathscr{W}}$};
        \node[inner sep=0pt, minimum size=1.2cm] (TB) at (15.3, 0.7) {$\mathscr{T}_{\mathscr{Z}}$};
        
        \node[fill=gray!10, draw=black, semithick, circle, inner sep=0pt, minimum size=0.85cm] (ai1) at (1, 2.75) {$w_{i}$};
        \node[fill=gray!10, draw=black, semithick, circle, inner sep=0pt, minimum size=0.85cm] (ai2) at (3, 2.75) {$w_{i^\prime}$};

        \node[fill=gray!10, draw=black, semithick, circle, inner sep=0pt, minimum size=0.85cm] (al2) at (12, 2.75) {$w_{l^\prime}$};
        \node[fill=gray!10, draw=black, semithick, circle, inner sep=0pt, minimum size=0.85cm] (al1) at (14, 2.75) {$w_{l}$};
        
        \node[fill=gray!10, draw=black, semithick, circle, inner sep=0pt, minimum size=0.85cm] (bj2) at (4, 0.25) {$z_{j^\prime}$};
        \node[fill=gray!10, draw=black, semithick, circle, inner sep=0pt, minimum size=0.85cm] (bj1) at (6, 0.25) {$z_{j}$};

        \node[fill=gray!10, draw=black, semithick, circle, inner sep=0pt, minimum size=0.85cm] (bk1) at (9, 0.25) {$z_{k}$};        
        \node[fill=gray!10, draw=black, semithick, circle, inner sep=0pt, minimum size=0.85cm] (bk2) at (11, 0.25) {$z_{k^\prime}$};

        \draw [line width=0.3mm, >=stealth, ->] (ai1) -- (bj1);
        \draw [line width=0.3mm, >=stealth, ->] (bk1) -- (al1);
        
        \draw [style=dashed, line width=0.3mm, >=stealth, ->] (ai2) -- (bj2);
        \draw [style=dashed, line width=0.3mm, >=stealth, ->] (bk2) -- (al2);
        
    \end{tikzpicture}
    \caption{An example of priority requests, where the demands for the customer $c\!=\!((w_i,z_j),(z_k,w_l))$ are displayed as full arrows, and for the customer $c^\prime\!=\!((w_{i^\prime},z_{j^\prime}),(z_{k^\prime},w_{l^\prime}))$ are displayed as dashed arrows.}
    \label{fig:compare}
\end{figure}
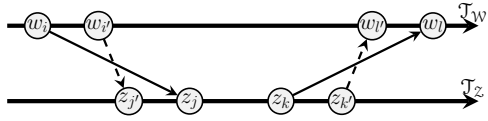

\begin{proof} Let $\mathcal{C}_{1}$, $c$ and $c^\prime$ be as stated. Note that, as customer $c$ is satisfied in $\mathcal{C}_{1}$, there is a car available at $w_i$ to fulfill the outbound demand, and there is a car available at $z_k$ to fulfill the return demand (see Fig.~\ref{fig:compare}). Since $w_{i} \leq w_{i^\prime}$ and $z_{k} \leq z_{k^\prime}$, we can exchange $c$ for $c^\prime$ in $\mathcal{C}_{1}$ to obtain $\mathcal{C}_{2}$. Let us now consider $\mathscr{T}^\prime\!=\!\{w_i, \dots, w_{i^\prime}\} \cup \{w_{l^\prime}, \dots, w_l\} \cup \{z_{j^\prime}, \dots, z_j\} \cup \{z_k, \dots, z_{k^\prime}\}$. Now, notice that there is the same amount of cars available at the vertices in $(\mathscr{T}_{\hspace*{-.2em}\mathscr{W}}\cup\mathscr{T}_{\hspace*{-.2em}\mathscr{Z}}) \setminus \mathscr{T}^\prime$ as before, and there is one more car available at the vertices in  $\mathscr{T}^\prime$. Thus, $\mathcal{C}_{2}$ is feasible.
\end{proof}

Therefore, for a pair $c$ and $c^\prime$ of customers as stated in Proposition~\ref{prop:inequality}, we can add (using a tie-breaking rule) the following constraint to ensure that the customer $c$ will be satisfied only if the customer $c^\prime$ is satisfied:
\begin{equation}
x_{\outbound{c}} \leq x_{\outbound{c^\prime}}. \label{eq7} 
\end{equation}

\begin{figure}[!h]

    \LARGE
    \centering

    \definecolor{c1}{rgb}{0.00, 0.50, 0.00} 
    \definecolor{c2}{rgb}{0.81, 0.09, 0.13} 
    \definecolor{c3}{rgb}{0.69, 0.40, 0.00} 
    \definecolor{c4}{rgb}{0.03, 0.27, 0.49} 
    \definecolor{c5}{rgb}{0.54, 0.17, 0.89} 
        
    \begin{subfigure}[b]{0.65\textwidth}
    \centering
    \begin{tikzpicture}[scale=.4, transform shape]
         
        \foreach \i in {0.25, 2.75} {
            \draw[ultra thick, >=stealth, ->] (0, \i) -- (26.0, \i);
        }
        
        \node[inner sep=0pt, minimum size=1.4cm] (TA) at (25.8, 3.2) {$\TA$};
        \node[inner sep=0pt, minimum size=1.2cm] (TB) at (25.8, 0.7) {$\TB$};
        
        \setcounter{count}{1}
        \foreach \i in {1, 2.5, 4, 5.5, 18.5, 20, 21.5, 23, 24.5} {
            \def\k{\the\value{count}}
            \node[fill=gray!10, draw=black, semithick, circle, inner sep=0pt, minimum size=0.85cm] (a\k) at (\i, 2.75) {$a_{\k}$};
            \stepcounter{count}
        }
        
        \setcounter{count}{1}
        \foreach \i in {6, 7.5, 9, 10.5, 12, 15, 16.5, 18} {
            \def\k{\the\value{count}}
            \node[fill=gray!10, draw=black, semithick, circle, inner sep=0pt, minimum size=0.85cm] (b\k) at (\i, 0.25) {$b_{\k}$};
            \stepcounter{count} 
        }
        
        \draw [c1!80, line width=0.3mm, >=stealth, ->] (a4) -- (b1);
        \draw [c1!80, line width=0.3mm, >=stealth, ->] (b8) -- (a5);
        \draw [c2!80, line width=0.3mm, >=stealth, ->] (a2) -- (b2);
        \draw [c2!80, line width=0.3mm, >=stealth, ->] (b7) -- (a6);
        \draw [c3!80, line width=0.3mm, >=stealth, ->] (a3) -- (b3);
        \draw [c3!80, line width=0.3mm, >=stealth, ->] (b6) -- (a7);
        \draw [c4!80, line width=0.3mm, >=stealth, ->] (a2) -- (b4);
        \draw [c4!80, line width=0.3mm, >=stealth, ->] (b6) -- (a8);
        \draw [c5!80, line width=0.3mm, >=stealth, ->] (a1) -- (b5);
        \draw [c5!80, line width=0.3mm, >=stealth, ->] (b6) -- (a9);

        \node[fill=white, draw=white, semithick, circle, inner sep=0pt, minimum size=0.85cm] (a) at (  0,  4.75) {};
        \node[fill=white, draw=white, semithick, circle, inner sep=0pt, minimum size=0.85cm] (e) at (.75, -1.75) {};

    \end{tikzpicture}
    \end{subfigure}
    \begin{subfigure}[b]{0.20\textwidth}
    \centering
    \begin{tikzpicture}[scale=.4, transform shape]
    
        \node[fill=c1!50, draw=black, semithick, circle, inner sep=0pt, minimum size=0.85cm] (a) at (  0,  4.75) {$c_{1}$};
        \node[fill=c2!50, draw=black, semithick, circle, inner sep=0pt, minimum size=0.85cm] (b) at ( -2,  2.75) {$c_{2}$};
        \node[fill=c3!50, draw=black, semithick, circle, inner sep=0pt, minimum size=0.85cm] (c) at (  1,  2.75) {$c_{3}$};
        \node[fill=c4!50, draw=black, semithick, circle, inner sep=0pt, minimum size=0.85cm] (d) at (-.5,  0.75) {$c_{4}$};
        \node[fill=c5!50, draw=black, semithick, circle, inner sep=0pt, minimum size=0.85cm] (e) at (.75, -1.75) {$c_{5}$};
        
        \draw [line width=0.3mm, >=stealth, ->] (a) -- (b);
        \draw [line width=0.3mm, >=stealth, ->] (a) -- (c);
        \draw [line width=0.3mm, >=stealth, ->] (a) -- (d);
        \draw [line width=0.3mm, >=stealth, ->] (a) to [out=1, in=30] (e);
        \draw [line width=0.3mm, >=stealth, ->] (b) -- (d);
        \draw [line width=0.3mm, >=stealth, ->] (b) to [out=250, in=180] (e);
        \draw [line width=0.3mm, >=stealth, ->] (c) -- (d);
        \draw [line width=0.3mm, >=stealth, ->] (c) -- (e);
        \draw [line width=0.3mm, >=stealth, ->] (d) -- (e);
   
    \end{tikzpicture}
    \end{subfigure}
    \begin{subfigure}[b]{0.10\textwidth}
    \centering
    \begin{tikzpicture}[scale=.4, transform shape]
    
        \node[fill=c1!50, draw=black, semithick, circle, inner sep=0pt, minimum size=0.85cm] (a) at (  0,  4.75) {$c_{1}$};
        \node[fill=c2!50, draw=black, semithick, circle, inner sep=0pt, minimum size=0.85cm] (b) at ( -2,  2.75) {$c_{2}$};
        \node[fill=c3!50, draw=black, semithick, circle, inner sep=0pt, minimum size=0.85cm] (c) at (  1,  2.75) {$c_{3}$};
        \node[fill=c4!50, draw=black, semithick, circle, inner sep=0pt, minimum size=0.85cm] (d) at (-.5,  0.75) {$c_{4}$};
        \node[fill=c5!50, draw=black, semithick, circle, inner sep=0pt, minimum size=0.85cm] (e) at (.75, -1.75) {$c_{5}$};
        
        \draw [line width=0.3mm, >=stealth, ->] (a) -- (b);
        \draw [line width=0.3mm, >=stealth, ->] (a) -- (c);
        \draw [line width=0.3mm, >=stealth, ->] (a) -- (d);
        \draw [line width=0.3mm, >=stealth, ->] (b) -- (d);
        \draw [line width=0.3mm, >=stealth, ->] (c) -- (d);
        \draw [line width=0.3mm, >=stealth, ->] (d) -- (e);
   
    \end{tikzpicture}
    \end{subfigure}
    
    \caption{An example of a directed acyclic graph interpreting the pairs of the of the customers' priority requests (on the left) and its transitive reduction (on the right).}
    \label{fig:transitive_reduction}
\end{figure}
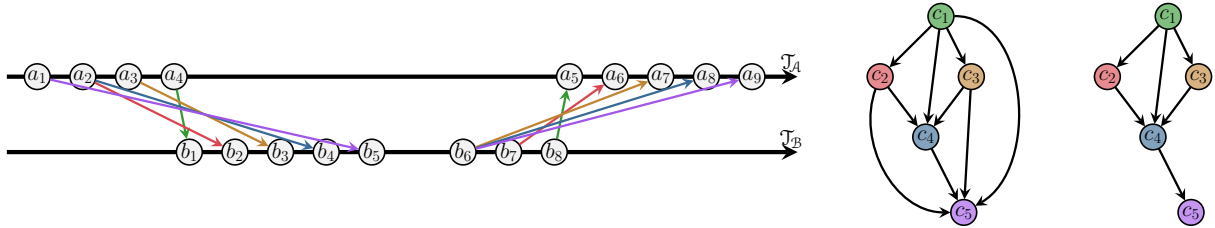

During preliminary tests, we observed that the model obtained by adding all possible constraints had a computational performance slightly worse than the model without these inequalities. To significantly improve the computational performance, we created a (disconnected) directed acyclic graph in which each arc represents the priority requests relationships, and we used a transitive reduction to get a minimum equivalent digraph that has the same transitivity closure and as few arcs as possible (see an example in Fig.~\ref{fig:transitive_reduction}). A description of the transitive reduction of a directed graph can be found in Aho et al.~\cite{Aho1972}.

We also observed that removing more arcs to obtain an arborescence forest (a directed acyclic graph with maximum in-degree one) after the transitive reduction leads to a significant gain in the computational time, proving optimality for more instances in less computing time. To ensure that the node has at most one parent node in the forest, we have chosen to keep as the parent node the one with the smallest work schedule. Here, the \emph{work schedule} is the difference between the start time of the outbound request and the end time of the return request. Therefore the remaining arcs of this forest indicate the constraints to be added to the formulation. We will call this strengthened model (CS2).

    \section{Matheuristic algorithms}
\label{sec:heuristics}

In this section, we propose a greedy randomized adaptive search procedure (GRASP), a variable neighborhood search (VNS), and a tabu search (TS) to find feasible solutions for the {\probInitials}. First, we start this section by discussing the difficulties in designing heuristics that produce high-quality solutions. Next, we present the main ingredients of the heuristic procedures, \ie, the neighborhood structures, the local search, and the proposed heuristic algorithms. Finally, we describe the implementation details of how we check if a solution is feasible or not.

\subsection{Downward closed analysis}

For several combinatorial optimization problems such as Knapsack and Maximum Independent Set, any feasible solution (and, thus, any optimal solution) can be constructed by starting with an empty solution and adding instance elements (such as items or vertices) iteratively in some order. Thus, in preliminary studies, we implemented constructive heuristics for the {\probInitials} which, unfortunately, led to low-quality solutions. 

Consider a family $\mathcal{S}$ of sets, we say that $\mathcal{S}$ is downward closed if and only if for any $S \in \mathcal{S}$ and any $T \subseteq S$ we have that $T \in \mathcal{S}$.
Next, we show that {\probInitials} does not have this property, which can explain why constructive heuristics have performed~poorly.

\begin{figure}[!h]

    \LARGE
    \centering
    
    \begin{tikzpicture}[scale=.4, transform shape]
        
        \centering
        
        \definecolor{c4}{rgb}{0.03, 0.27, 0.49}
        \definecolor{c2}{rgb}{0.81, 0.09, 0.13}
        \definecolor{c3}{rgb}{0.69, 0.40, 0.00}
        \definecolor{c1}{rgb}{0.00, 0.50, 0.00}

        \node[fill=gray!10, draw=black, semithick, circle, inner sep=0pt, minimum size=0.85cm] (s) at (-.5,1.5) {$s$};
        \node[fill=gray!10, draw=black, semithick, circle, inner sep=0pt, minimum size=0.85cm] (t) at (16,1.5) {$t$};
        
        \setcounter{count}{1}
        
        \foreach \i in {1, 5.25, 9.5, 14} {
            \def\k{\the\value{count}}
            \node[fill=gray!10, draw=black, semithick, circle, inner sep=0pt, minimum size=0.85cm] (a\k) at (\i, 2.75) {$a_{\k}$};
            \stepcounter{count}
        }
        
        \setcounter{count}{1}
        
        \foreach \i in {1.5, 5.75, 10, 14.5} {
            \def\k{\the\value{count}}
            \node[fill=gray!10, draw=black, semithick, circle, inner sep=0pt, minimum size=0.85cm] (b\k) at (\i, 0.25) {$b_{\k}$};
            \stepcounter{count} 
        }
    
        \foreach \i [remember=\i as \lasti (initially 1)]  in {2, ..., 4} {
            \draw [style=dashed, line width=0.275mm, >=stealth, ->] (a\lasti) -- (a\i);
            \draw [style=dashed, line width=0.275mm, >=stealth, ->] (b\lasti) -- (b\i);
        }
        
        \draw [style=dotted, line width=0.4mm, >=stealth, ->] (s) -- (a1) node[midway, above]{$1$~~~~}; 
        \draw [style=dotted, line width=0.4mm, >=stealth, ->] (s) -- (b1) node[midway, below]{$1$~~}; 
        \draw [style=dashed, line width=0.275mm, >=stealth, ->] (a4) -- (t); 
        \draw [style=dashed, line width=0.275mm, >=stealth, ->] (b4) -- (t);
     
        \draw [c3, line width=0.3mm, >=stealth, ->] 
        (a1) -- (b1) node[midway, above]{\textcolor{black}{~~~\,$o_{3}$}};
        \draw [c3, line width=0.3mm, >=stealth, ->] 
        (b2) -- (a4) node[near start, above]{\textcolor{black}{$r_{3}$\,}};
        \draw [c2, line width=0.3mm, >=stealth, ->] 
        (a2) -- (b2) node[midway, above]{\textcolor{black}{$o_{2}$~~~~}};
        \draw [c2, line width=0.3mm, >=stealth, ->] 
        (b3) -- (a4) node[near start, xshift=-.08cm, yshift=0cm, above]{\textcolor{black}{$r_{2}$\,}};
        \draw [c4, line width=0.3mm, >=stealth, ->] 
        (b1) -- (a2) node[midway, xshift=0.05cm, yshift=-0.1cm, above]{\textcolor{black}{$o_{4}$~~~\,}};
        \draw [c4, line width=0.3mm, >=stealth, ->] 
        (a3) -- (b3) node[midway, above]{\textcolor{black}{$r_{4}$~~~~}};
        \draw [c1, line width=0.3mm, >=stealth, ->] 
        (b1) -- (a3) node[near start, below]{\textcolor{black}{$o_{1}$}};
        \draw [c1, line width=0.3mm, >=stealth, ->] 
        (a4) -- (b4) node[near end, xshift=-0.10cm, yshift=0cm, above]{\textcolor{black}{$r_{1}$~~~}};
        
    \end{tikzpicture}
    \caption{An example in which all customers must be satisfied simultaneously.}
    \label{fig:sample3}
\end{figure}
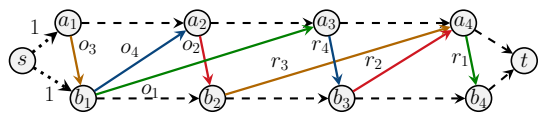

Figure~\ref{fig:sample3} illustrates an instance where all customers can be satisfied simultaneously. Suppose that a feasible solution is constructed, iteratively, by adding customers over the permutation $\langle c_3, c_4, c_2, c_1 \rangle$. In this case, only customers $c_3$ and~$c_4$ can be satisfied. Note that trying to satisfy customers $c_3$, $c_4$ and $c_2$ will result in a car missing at $a_\textnormal{3}$. The same applies when trying to satisfy customer $c_1$. Here, any subset of three customers is an infeasible solution, thus any ordering of customers cannot lead to the optimal solution when considering a constructive heuristic that adds one customer per iteration.

This observation led to designing heuristics that removes customers from an infeasible solution until obtaining a feasible solution that performed better than iteratively adding customers. Also, it led to the design of neighborhood structures that adds more than one customer at a time.

\subsection{Neighborhood structures}

We propose neighborhood structures defined on the space of feasible solutions, \ie, only feasible moves are accepted. The following neighborhood structures were implemented and used in shaking and improvement procedures.

\begin{itemize}
    \item $\mathcal{N}_{1}$ -- \emph{increase}: The neighbors of a solution are the solutions obtained by adding one (previously) unsatisfied customer into the solution.
    \item $\mathcal{N}_{2}$ -- \emph{double increase}: The neighbors of a solution are the solutions obtained adding two unsatisfied customers into the solution. If one of the customers can be added to the solution while maintaining feasibility, the pair is disregarded. 
    \item $\mathcal{N}_{3}$ -- \emph{triple increase}: The neighbors of a solution are the solutions obtained by adding three unsatisfied customers into the solution.
    If up to two of the customers can be added to the solution while maintaining feasibility, the triple is disregarded. 
    \item $\mathcal{N}_{4}$ -- \emph{decrease}: The neighbors of a solution are the solutions obtained by removing one satisfied customer from the solution.
    \item $\mathcal{N}_{5}$ -- \emph{double decrease}: The neighbors of a solution are the solutions obtained by removing two satisfied customers from the solution.
    If one of the customers can be removed from the solution while maintaining feasibility, the pair is disregarded. 
    \item $\mathcal{N}_{6}$ -- \emph{triple decrease}: The neighbors of a solution are the solutions obtained by removing three satisfied customers from the solution.
    If up to two of the customers can be removed from the solution while maintaining feasibility, the triple is disregarded. 
    \item $\mathcal{N}_{7}$ -- \emph{exchange one for one}: The neighbors of a solution are the solutions obtained by exchanging one satisfied customer from the solution for one unsatisfied customer.
    \item $\mathcal{N}_{8}$ -- \emph{exchange one for two}: The neighbors of a solution are the solutions obtained by exchanging one satisfied customer from the solution for two unsatisfied customers.
\end{itemize}

Also, for $1 \leq i \leq 8$, we denote the set of neighbors of a feasible solution $\mathcal{C}^\prime \subseteq \mathcal{C}$ in neighborhood~$\mathcal{N}_{i}$ by~$\mathcal{N}_{i}(\mathcal{C}^\prime)$.

\subsection{Local search}
\label{subsec:local_search}

The local search heuristic (presented in Algorithm~\ref{alg:local_search}) repeatedly improves a solution $\mathcal{C}^\prime \subseteq \mathcal{C}$ until $\mathcal{N}_{2}(\mathcal{C}^\prime) \cup \mathcal{N}_{1}(\mathcal{C}^\prime) \cup \mathcal{N}_{8}(\mathcal{C}^\prime)$ is empty and, thus, until $\mathcal{C}^\prime$ is locally optimal for these neighborhood structures. 

The heuristic first improves the solution as much as possible using~$\mathcal{N}_{2}$. Afterward, the heuristic makes a single improvement using either $\mathcal{N}_{1}$, if $\mathcal{N}_{1}$ is not empty, or $\mathcal{N}_{8}$, otherwise. If the solution is not locally optimal, the process is repeated starting from $\mathcal{N}_{2}$. At each point, when considering a neighborhood structure $\mathcal{N}_{i}$, if $\mathcal{N}_{i}$ is non-empty, $\mathcal{C}^\prime$ is replaced by an uniformly randomly chosen solution from $\mathcal{N}_{i}(\mathcal{C}^\prime)$.

\begin{algorithm}[!h]
    \footnotesize   
    \caption{Local Search} 
    \label{alg:local_search}
    \begin{algorithmic}[1]
        \Repeat 
            \While {$\mathcal{N}_{2}(\mathcal{C}^\prime)$ is non-empty}
                \State Randomly choose a neighbor $\mathcal{C}^{\prime\prime}$ from the neighboring of $\mathcal{N}_{2}(\mathcal{C}^\prime)$ 
                \State $\mathcal{C}^\prime \gets \mathcal{C}^{\prime\prime}$
            \EndWhile
            \State Set $k$ to 1 if  $\mathcal{N}_{1}(\mathcal{C}^\prime)$ is non-empty; otherwise, set $k$ to $8$
            \If {$\mathcal{N}_{k}(\mathcal{C}^\prime)$ is non-empty}
                \State Randomly choose a neighbor $\mathcal{C}^{\prime\prime}$ from the neighboring of $\mathcal{N}_{k}(\mathcal{C}^\prime)$
                \State $\mathcal{C}^\prime \gets \mathcal{C}^{\prime\prime}$
            \EndIf
        \Until {$\mathcal{C}^\prime$ is not locally optimum}
        \State \Return $\mathcal{C}^\prime$
    \end{algorithmic} 
\end{algorithm}

\subsection{Greedy randomized adaptive search procedure}

The Greedy Randomized Adaptive Search Procedure (GRASP) is a multi-start metaheuristic method composed of two phases, a construction phase and a local search phase. A description of GRASP can be found in Feo and Resende~\cite{Feo1995}. In our proposed GRASP, at each iteration, a constructed initial solution is obtained through solving several linear relaxation of (CS2) with fixed values for some variables in order to obtain a high-quality solution for the {\probInitials}. For simplicity, we will denote the linear relaxation of~(CS2) by $\mathrm{(CS2)_R}$.
\begin{algorithm}[!h]
    \footnotesize   
    \caption{Greedy Randomized Construction} 
    \label{alg:greedy_randomized_construction}
    \begin{algorithmic}[1]
        \State Compute an optimal solution $x^*$ for $\mathrm{(CS2)_R}$
        \State Define CL as all customer $c \in \mathcal{C}$, such that $x^*_{d_c} \geq 1/2$
        \While {is not possible satisfy all customer in CL}
            \For {all candidate customer $c$ in CL}
                \State Define $eval(c)$ as 
                the optimal value of $\mathrm{(CS2)_R}$ restricted to $x_{c'} = 0$ for all $c' \notin \mathrm{CL} \setminus \{c\}$
            \EndFor
            \State Define $\Delta_{min}$ as $\min\{eval(c)\colon c \in \mathrm{CL}\}$ and $\Delta_{max}$ as $\max\{eval(c)\colon c \in \mathrm{CL}\}$
            \State Define RCL as $\{c \in \mathrm{CL}\colon eval(c) \geq \Delta_{min} + \alpha(\Delta_{max} - \Delta_{min})\}$
            \State Randomly choose a customer $c$ in RCL
            \State Remove $c$ from CL
            \State Define $x^*$ as an optimal solution found in line 5 when considering customer $c$
            \State Remove from CL each customer $c^\prime$ where $x^*_{d_{c^\prime}} < 1/2$
        \EndWhile
        \State \Return CL
    \end{algorithmic} 
\end{algorithm}

Algorithm~\ref{alg:greedy_randomized_construction} presents the construction phase. Initially, the candidate list CL is created by selecting each customer $c \in \mathcal{C}$, where $x^*_{d_c} \geq 1/2$ from an optimal solution $x^*$ of $\mathrm{(CS2)_R}$ (lines~1 and~2). 
For every candidate customer $c$ in CL, the evaluation value of $c$ is defined by measuring the benefit of removing that candidate in the current (integer infeasible) solution. For this,~$\mathrm{(CS2)_R}$ is solved considering all candidate customers in CL, except for the candidate customer which will be evaluated (lines 4--6), and we define the evaluation value as the optimal objective value obtained.

The restricted candidate list RCL is, then, composed of each candidate customer in CL whose has the evaluated value at least $\Delta_{min} + \alpha(\Delta_{max} - \Delta_{min})$, where~$\Delta_{min}$ and $\Delta_{max}$ are the smallest and the largest evaluated value at the current iteration, respectively (lines~7 and~8). The parameter $\alpha \in [0, 1]$ controls the amounts of greediness and randomness in the construction phase. 

One candidate customer $c$ is selected uniformly at random from RCL to be removed from~CL. As removing only one candidate customer per iteration leads to a very slow algorithm, we also remove every candidate customer $c'$ that has $x^*_{d_{c'}} < 1/2$ in the optimal solution found for $c$ (lines 9--12). In this way, we aggressively drive to a feasible solution. When it is possible simultaneously to satisfy all the remaining candidate customers in CL, the construction phase ends and returns CL as a solution.

In the second phase, Algorithm~\ref{alg:local_search} is used to improve the constructed initial solution. The algorithm will repeat these steps until the computation time exceeds a maximum time value (termination criteria), and the best solution found over all GRASP iterations is returned as the result.

\subsection{Variable neighborhood search}

The Variable Neighborhood Search (VNS) is also a two-phase metaheuristic. It has a perturbation phase (a shaking mechanism) and an ascent phase (for maximization problems), based upon systematic changes of neighborhood structures in both phases. A description of~VNS can be found in Hansen and Mladenovi{\'c}~\cite{Hansen2001}. 

In our proposed VNS algorithm, presented in Algorithm~\ref{alg:vns}, an initial solution is obtained through the same construction procedure used in the GRASP, but with parameter $\alpha = 1$ (meaning that procedure is purely greedy).
\begin{algorithm}[!h]
    \footnotesize   
    \caption{Variable Neighborhood Search} 
    \label{alg:vns}
    \begin{algorithmic}[1]
        \State Build an initial solution $\mathcal{C}^\prime$ with a greedy constructive procedure
        \While {time limit is not met}
            \State Define $\mathcal{S} = \langle 7, 4, 5, 6\rangle$ as the sequential neighborhood change
            \While {$\mathcal{S}$ is not empty}
                \State Set $k$ to be the first element in $\mathcal{S}$
                \State Randomly choose $\mathcal{C}^{\prime\prime}$ from $\mathcal{N}_{k}(\mathcal{C}^\prime)$ 
                \State Apply a local search on $\mathcal{C}^{\prime\prime}$, obtaining the solution $\mathcal{C}^{\prime\prime\prime}$
                \If {$\mathcal{C}^{\prime\prime\prime}$ is better than $\mathcal{C}^\prime$}
                    \State $\mathcal{C}^\prime \gets \mathcal{C}^{\prime\prime\prime}$
                    \State Restore $\mathcal{S}$ to the initial sequential neighborhood change
                 \Else
                    \State Remove $k$ from $\mathcal{S}$
                \EndIf
            \EndWhile
        \EndWhile
        \State \Return $\mathcal{C}^\prime$
    \end{algorithmic} 
\end{algorithm}

We choose to use the neighborhood structures $\mathcal{N}_{7}$, $\mathcal{N}_{4}$, $\mathcal{N}_{5}$ and $\mathcal{N}_{6}$, in this order, in the shaking mechanism. Once an initial solution $\mathcal{C}^\prime \subseteq \mathcal{C}$ is constructed, the solution $\mathcal{C}^\prime$ is perturbed by randomly choosing a solution $\mathcal{C}^{\prime\prime}$ from the neighboring of $\mathcal{N}_{7}(\mathcal{C}^\prime)$ (by setting~${k = 7}$), and, then, the proposed local search is applied to improve this solution, obtaining a new solution $\mathcal{C}^{\prime\prime\prime}$ (lines 5--7). 

If any improvement is obtained, the solution $\mathcal{C}^{\prime}$ is replaced by the solution $\mathcal{C}^{\prime\prime\prime}$, and the search is configured to restart within the first neighborhood structure (lines 8--10). Otherwise, the next neighborhood structure is considered in the next iteration, by removing~$k$ from $\mathcal{S}$ (line 12), which performs a neighborhood change. 

The VNS algorithm repeats these steps until the computation time exceeds a time limit, and the current solution $\mathcal{C}^{\prime}$ is returned as the result.

\subsection{Tabu search}

The Tabu Search (TS) is a metaheuristic that guides a local search to explore the solution space beyond the highest peaks (local optimality) by prohibiting already visited solutions. An introduction to the fundamental principles of TS can be found in Glover~\cite{Glover1989,Glover1990}.

We propose a simplified version of TS for the {\probInitials}. Our TS algorithm, presented in Algorithm~\ref{alg:tabu_search}, starts from an initial solution $\mathcal{C}^\prime \subseteq \mathcal{C}$ provided by the same constructive procedure used in the VNS algorithm. Then, at any iteration, the search sequentially explores the proposed neighborhood structures, except for the neighborhood structures~$\mathcal{N}_{5}$ and~$\mathcal{N}_{6}$. Specifically, it explores the neighborhoods by selecting a random solution $\mathcal{C}^{\prime\prime}$ from $\mathcal{N}(\mathcal{C}^{\prime})$ (line 7). If the solution $\mathcal{C}^{\prime\prime}$ is a non-tabu solution, then the solution~$\mathcal{C}^{\prime}$ is replaced by the solution $\mathcal{C}^{\prime\prime}$ and a tabu list, containing the last visited solutions, is updated (lines 8--10). The solution $\mathcal{C}^{\prime\prime}$ will remain in the tabu list for a number of iterations obtaining new tabu solutions (as first in, first out queueing). If the tabu list is full, then old solution is replaced.
\begin{algorithm}[!h]
    \footnotesize   
    \caption{Tabu Search} 
    \label{alg:tabu_search}
    \begin{algorithmic}[1]
        \State Let $\mathcal{N} = \mathcal{N}_{1}\circ\mathcal{N}_{2}\circ\mathcal{N}_{3}\circ\mathcal{N}_{4}\circ\mathcal{N}_{7}\circ\mathcal{N}_{8}$ be an aggregated neighborhood structure
        \State Let $l$ be the desired maximum tabu list length
        \State Build an initial solution $\mathcal{C}^\prime$ with a greedy constructive procedure
        \State Update the incumbent solution $\hat{\mathcal{C}}$ to be $\mathcal{C}^\prime$ 
        \State Initialize the tabu list of maximum length $l$ to be empty
        \While {time limit is not met}
            \State Randomly choose $\mathcal{C}^{\prime\prime}$ from $\mathcal{N}(\mathcal{C}^\prime)$ 
            \If {$\mathcal{C}^{\prime\prime}$ is a non-tabu solution}
                \State $\mathcal{C}^\prime \gets \mathcal{C}^{\prime\prime}$
                \State Add $\mathcal{C}^{\prime\prime}$ to the tabu list removing the oldest solution from the tabu list if necessary
            \EndIf
            \If {$\mathcal{C}^{\prime\prime}$ is better than $\hat{\mathcal{C}}$}
                \State $\hat{\mathcal{C}} \gets \mathcal{C}^{\prime\prime}$
             \EndIf
        \EndWhile
        \State \Return $\hat{\mathcal{C}}$
    \end{algorithmic} 
\end{algorithm}

In our implementation, we designate the \emph{tabu tenure} as a parameter, between $0$ and $1$, which determines the relative tabu list length. The maximum tabu list length is defined by setting $l$ as the tabu tenure times the number of customers in $\mathcal{C}$ rounded down. 

When the exploration ends by time exceeding a maximum time value, the incumbent solution $\hat{\mathcal{C}}$ is returned as the result.

\subsection{Fast feasibility check}

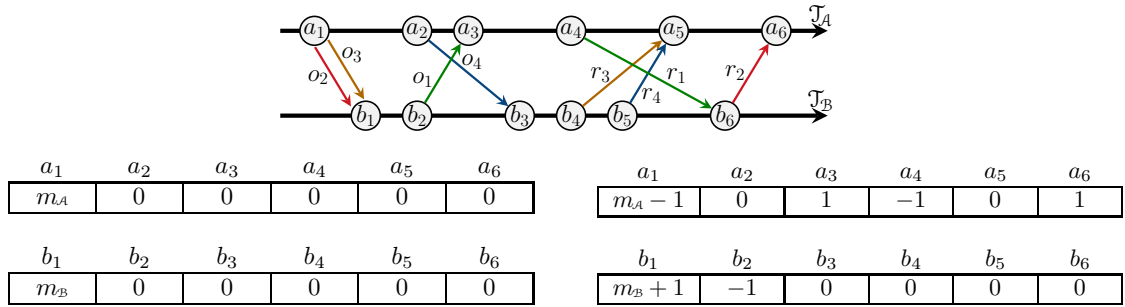
\begin{figure}[ht]
    
    \LARGE
    \centering
    
    \begin{tikzpicture}[scale=.45, transform shape]
    
        \centering
    
        \definecolor{c4}{rgb}{0.03, 0.27, 0.49}
        \definecolor{c2}{rgb}{0.81, 0.09, 0.13}
        \definecolor{c3}{rgb}{0.69, 0.40, 0.00}
        \definecolor{c1}{rgb}{0.00, 0.50, 0.00}
    
        \foreach \i in {0.25, 2.75} {
            \draw[ultra thick, >=stealth, ->] (0, \i) -- (16.0, \i);
        }
        
        \node[inner sep=0pt, minimum size=1.4cm] (TA) at (15.8, 3.2) {$\TA$};
        \node[inner sep=0pt, minimum size=1.2cm] (TB) at (15.8, 0.7) {$\TB$};
        
        \setcounter{count}{1}
        \foreach \i in {1, 4, 5.5, 8.5, 11.5, 14.5} {
            \def\k{\the\value{count}}
            \node[fill=gray!10, draw=black, semithick, circle, inner sep=0pt, minimum size=0.85cm] (a\k) at (\i, 2.75) {$a_{\k}$};
            \stepcounter{count}
        }
        
        \setcounter{count}{1}
        \foreach \i in {2.5, 4, 7, 8.5, 10, 13} {
            \def\k{\the\value{count}}
            \node[fill=gray!10, draw=black, semithick, circle, inner sep=0pt, minimum size=0.85cm] (b\k) at (\i, 0.25) {$b_{\k}$};
            \stepcounter{count} 
        }
    
        \draw [c3, transform canvas={xshift=0.25ex,yshift=0.15ex}, line width=0.3mm, >=stealth, ->] 
        (a1) -- (b1) node[midway, above]{\textcolor{black}{~\,$o_{3}$}};
        \draw [c3, line width=0.3mm, >=stealth, ->] 
        (b4) -- (a5) node[near start, above]{\textcolor{black}{$r_{3}$\,}};
        \draw [c2, transform canvas={xshift=-.3ex,yshift=-.15ex}, line width=0.3mm, >=stealth, ->] 
        (a1) -- (b1) node[near start, below]{\textcolor{black}{$o_{2}$\,~}};
        \draw [c2, line width=0.3mm, >=stealth, ->] 
        (b6) -- (a6) node[near start, xshift=-.08cm, yshift=0cm, above]{\textcolor{black}{$r_{2}$~}};
        \draw [c4, line width=0.3mm, >=stealth, ->] 
        (a2) -- (b3) node[midway, xshift=0.05cm, yshift=-0.1cm, above]{\textcolor{black}{\,$o_{4}$}};
        \draw [c4, line width=0.3mm, >=stealth, ->] 
        (b5) -- (a5) node[midway, xshift=-0.18cm, yshift=-0.2cm, below]{\textcolor{black}{~~\,$r_{4}$}};
        \draw [c1, line width=0.3mm, >=stealth, ->] 
        (b2) -- (a3) node[near start, xshift=-0.15cm, yshift=-0.2cm, above]{\textcolor{black}{$o_{1}$\,~}};
        \draw [c1, line width=0.3mm, >=stealth, ->] 
        (a4) -- (b6) node[near end, xshift=-0.10cm, yshift=0cm, above]{\textcolor{black}{$r_{1}$}};
    
    \end{tikzpicture}
    
    \vspace*{0.25cm}
    \mbox{
        \footnotesize
        \begin{subfigure}[b]{0.42\textwidth}
            \centering
            \resizebox{\columnwidth}{!}{%
                \begin{tabular}{cccccc}
                ~~$a_\textnormal{1}$~~ & ~~$a_\textnormal{2}$~~ & ~~$a_\textnormal{3}$~~ & ~~$a_\textnormal{4}$~~ & ~~$a_\textnormal{5}$~~ & ~~$a_\textnormal{6}$~~ \\
                \hline
                \multicolumn{1}{|c|}{$\nA$} & 
                \multicolumn{1}{c|}{$0$} & 
                \multicolumn{1}{c|}{$0$} & 
                \multicolumn{1}{c|}{$0$} & 
                \multicolumn{1}{c|}{$0$} & 
                \multicolumn{1}{c|}{$0$} \\ 
                \hline \\
                $b_\textnormal{1}$ & $b_\textnormal{2}$ & $b_\textnormal{3}$ & $b_\textnormal{4}$ & $b_\textnormal{5}$ & $b_\textnormal{6}$ \\ 
                \hline
                \multicolumn{1}{|c|}{$\nB$} & 
                \multicolumn{1}{c|}{$0$} & 
                \multicolumn{1}{c|}{$0$} & 
                \multicolumn{1}{c|}{$0$} & 
                \multicolumn{1}{c|}{$0$} & 
                \multicolumn{1}{c|}{$0$} \\ 
                \hline
                \end{tabular}%
            }
            \label{fig:subfig4}
        \end{subfigure}
        \hspace*{0.6cm}
        \begin{subfigure}[b]{0.42\textwidth}
            \centering
            \resizebox{\columnwidth}{!}{%
                \begin{tabular}{cccccc}
                ~~$a_\textnormal{1}$~~ & ~~$a_\textnormal{2}$~~ & ~~$a_\textnormal{3}$~~ & ~~$a_\textnormal{4}$~~ & ~~$a_\textnormal{5}$~~ & ~~$a_\textnormal{6}$~~ \\ 
                \hline
                \multicolumn{1}{|c|}{$\nA -1$} & 
                \multicolumn{1}{c|}{$0$} & 
                \multicolumn{1}{c|}{$1$} & 
                \multicolumn{1}{c|}{$-1$} & 
                \multicolumn{1}{c|}{$0$} & 
                \multicolumn{1}{c|}{$1$} \\ 
                \hline \\
                $b_\textnormal{1}$ & $b_\textnormal{2}$ & $b_\textnormal{3}$ & $b_\textnormal{4}$ & $b_\textnormal{5}$ & $b_\textnormal{6}$ \\ 
                \hline
                \multicolumn{1}{|c|}{$\nB +1$} & 
                \multicolumn{1}{c|}{$-1$} & 
                \multicolumn{1}{c|}{$0$} & 
                \multicolumn{1}{c|}{$0$} & 
                \multicolumn{1}{c|}{$0$} & 
                \multicolumn{1}{c|}{0} \\ 
                \hline
                \end{tabular}%
            }
            \label{fig:subfig5}
        \end{subfigure}
    }
    \caption{Examples for the exact configurations of the incremental displacement vectors. On the left, the configuration of an empty solution, and, on the right, the configuration of a solution where the customers~$c_1$ and~$c_2$ are satisfied.
    } 
    
    \label{fig:sample2}
\end{figure}

In this section, we present an efficient mechanism for checking if a solution is feasible or infeasible. For this, we use the prefix sum to determine the number of cars available, and the minimum prefix sum to determine the smallest number of cars available in a given range of vertices of a station. In this case, we use two incremental displacement vectors (see Fig.~\ref{fig:sample2} for examples), where each vector is associated with a station. 
 
Following the example, note that at the first position of each incremental displacement vector is initialized respectively with the fleet sizes $\nA$ and $\nB$, and the remaining positions are initialized with $0$. Furthermore, an operation to insert an unsatisfied customer into the current solution decreases the incremental displacement vector by a unit value at the start-points and increases it by a unit value at the end-points. Inversely, an operation to remove a customer of the current solution increase of a unit value at the start-points and decrease of a unit value at the end-points.

To allow for a fast evaluation of moves in the neighborhoods, we implemented a segment tree data structure to represent each incremental displacement vector. This data structure is critical for the heuristics' performance because it can compute the prefix sum and the minimum prefix sum of the incremental displacement vector in logarithmic time. 

We can check for a given customer if there is a car available before inserting them in the current solution using the prefix sum for each demand's start-point. And we can check if the current solution will remain feasible using the minimum prefix sum in the range that will have one car less available, combined with the number of cars available at the point before the range considered.
    \section{Computational results and discussion}
\label{sec:results}

All computational tests were performed on a machine with an Intel Xeon X3430 2.4GHz processor and 8GB of RAM, under Ubuntu 18.04 LTS environment. The approaches in the previous section have been implemented in the {\CC} language (code was compiled with optimization flag -O3 using GCC version 4.8.4), and Gurobi Optimizer version 8.0.1 (using a single thread).

Based on empirical observations, we consider for GRASP and TS, respectively, the value of $\alpha$ equal to $0.8$ and the value of the tabu tenure as $0.046$.

\subsection{Benchmark instances} 

For the computational experiments, we created three groups of random instances, where every group has sets of $100$ instances with $1000$, $2500$, and $5000$ customers. All instances and the source code are publicly available at \textsf{\hyperlink{https://gitlab.com/welverton/car-sharing_2020}{https://gitlab.com/welverton/car-sharing\_2020}} to facilitate future comparisons. These instances were generated as described below, considering a 1-day planning horizon.
\begin{itemize}

	\item \texttt{st}: In this group, every demand for driving is fulfilled within $15$ to $60$ minutes. Each demand is generated by setting the time points $t_1$ and $t_2$ randomly according to a uniform distribution, and time for driving values $d_1$ and $d_2$ randomly drawn from a uniform distribution over the integers from $15$ to $60$, such that $t_1+d_1 < t_2$ and $t_2+d_2 \leq 1440$. In this case, request for outbound starts at time $t_1$ and ends at time $t_1+d_1$, and request for return starts at time $t_2$ and ends at time $t_2+d_2$.
	
	\item \texttt{ft}: Every customer has exactly the same time for each driving demand, fixed within $15$ to $45$ minutes. First, it is fixed a time value for driving $d$, randomly drawn from a uniform distribution over the integers from $15$ to $45$. Next, each demand is generated by setting the time points $t_1$ and $t_2$ randomly according to a uniform distribution, such that $t_1+d < t_2$ and $t_2+d \leq 1440$. Thus, request for outbound starts at time $t_1$ and ends at time $t_1+d$, and request for return starts at time $t_2$ and ends at time $t_2+d$. 
	
	\item \texttt{fc}: Every customer has exactly the same time for each driving demand, fixed within $15$ to $45$ minutes, and has the same working time, fixed within $60$ to $240$ minutes. First, it is fixed a time value for driving $d$ and a working time value $w$, randomly drawn from a uniform distribution over the integers from $15$ to $45$ and from $60$ to $240$, respectively. Next, each demand is generated by setting the time point $t_1$ randomly according to a uniform distribution, and setting $t_2$ to $t_1+d+w$, such that $t_2+d \leq 1440$. Equal to the previous group, request for outbound starts at time $t_1$ and ends at time $t_1+d$, and request for return starts at time $t_2$ and ends at time $t_2+d$. 
\end{itemize}

For each customer, the direction of the first demand is chosen uniformly at random and the direction of the second demand is assigned oppositely. Every station has exactly $10$ cars initially available.

\subsection{MIP model experiments}

In this section, we report and discuss the results obtained by means of the {\probInitials} formulation. We consider the two previously mentioned MIP models~(CS1) and~(CS2) where each pair of variable $x_{\outbound{c}}$ and $x_{\return{c}}$ was replaced by a single variable $x_{d_c}$ (as observed in Section~\ref{sec:mip}).

Table~\ref{tab:MIPmodels-avg-results} summarizes the results with computing time limited to $10$ minutes. In this table, columns \emph{\small \#Opt} report, per instance set, the number of instances solved to (proven) optimality. In addition, columns \emph{\small Avg. satisfied customers}, \emph{\small Avg. gap (\%)} and \emph{\small Avg. constraints~\eqref{eq7}} report the average number of satisfied customers, the average relative gap (computed as \mbox{\small $100 \cdot (\UB-\LB)/\UB$}, where {\small $\LB$} is the lower bound and {\small $\UB$} is the upper bound), and the average number of constraints \eqref{eq7} added to (CS2), respectively. Values after $\pm$ denote the standard deviation.
\begin{table}[h!]
\centering
\caption{Aggregated MIP models results results for each instances set. In {\bf boldface}, the best number of problem instances solved to proven optimality and the best average of relative gap value (when there is no tie).}
\label{tab:MIPmodels-avg-results}
\resizebox{\textwidth}{!}{%
\begin{tabular}{cllrrrrrrrr}
\hline
\hline \vspace*{-0.3cm}\\
 &                     &  & \multicolumn{3}{c}{(CS1)}                                   &  & \multicolumn{4}{c}{(CS2)} \\ 
                            \cline{4-6}                                                      \cline{8-11} 
 & Instance set        &  & \#Opt  & Avg. satisfied customers &           Avg. gap (\%) &  &   \#Opt & Avg. satisfied customers &            Avg. gap (\%) & Avg. constraints \eqref{eq7} \\
\hline
 & \texttt{st}-$n$1000 &  &    75  &      403.99 $\pm$   4.13 &     0.064 $\pm$  0.114  &  &     75  &      403.99 $\pm$   4.15 &  \bf{0.062 $\pm$  0.108} &   35.53 $\pm$  5.37 \\
 & \texttt{st}-$n$2500 &  &     0  &      488.00 $\pm$   4.47 & \bf{0.514 $\pm$  0.125} &  &      0  &      488.00 $\pm$   4.48 &      0.522 $\pm$  0.134  &  204.54 $\pm$ 13.03 \\
 & \texttt{st}-$n$5000 &  &     0  &      549.17 $\pm$   3.83 & \bf{0.723 $\pm$  0.134} &  &      0  &      549.00 $\pm$   3.96 &      0.756 $\pm$  0.162  &  727.69 $\pm$ 25.33 \\
 & \texttt{ft}-$n$1000 &  &    83  &      478.47 $\pm$ 132.32 &     0.048 $\pm$  0.108  &  & \bf{85} &      478.46 $\pm$ 132.32 &  \bf{0.046 $\pm$  0.112} &    0.21 $\pm$  0.43 \\
 & \texttt{ft}-$n$2500 &  & \bf{3} &      512.85 $\pm$ 165.42 & \bf{1.507 $\pm$  0.436} &  &      0  &      512.80 $\pm$ 165.39 &      1.517 $\pm$  0.360  &    1.57 $\pm$  1.22 \\
 & \texttt{ft}-$n$5000 &  & \bf{6} &      239.57 $\pm$ 195.42 &    41.639 $\pm$ 46.420  &  &      5  &      254.82 $\pm$ 199.78 & \bf{39.065 $\pm$ 45.984} &    6.95 $\pm$  2.53 \\
 & \texttt{fc}-$n$1000 &  &    94  &      448.64 $\pm$ 121.55 &     0.017 $\pm$  0.070  &  & \bf{95} &      448.65 $\pm$ 121.53 &  \bf{0.015 $\pm$  0.066} &  177.79 $\pm$ 12.62 \\
 & \texttt{fc}-$n$2500 &  &    58  &      492.07 $\pm$ 154.76 & \bf{0.216 $\pm$  0.323} &  & \bf{59} &      492.00 $\pm$ 154.79 &      0.230 $\pm$  0.348  &  929.40 $\pm$ 36.16 \\
 & \texttt{fc}-$n$5000 &  &    54  &      501.95 $\pm$ 161.62 & \bf{0.337 $\pm$  0.538} &  & \bf{56} &      501.97 $\pm$ 161.66 &      0.340 $\pm$  0.528  & 2859.84 $\pm$ 68.22\vspace*{-0.3cm} \\ \\
\hline
\hline
\multicolumn{11}{l}{\small \emph{Note.} Avg., average; \#Opt, number of optimal solutions.}
\end{tabular}%
}
\end{table}

In terms of relative gaps, the results showed that both models were able to obtain good bounds for most of the instances, except for the instances of the group \texttt{ft} with $5000$ customers (models (CS1) and (CS2) did not found a feasible solution in $36$ and $35$ instances, respectively). 

In another experiment, we focused on instance set \texttt{st}-$n$1000 and let both models run until they could find an optimal solution and prove its optimality for all instances. 

We compared the root relaxation objective value and the optimal solution value and we observed that both models obtained, on average, $0.295\%$ of gap and $0.073$ of deviation from the mean. 

Based on the observation that the models obtain good upper bounds in the root node, we notice that it is more difficult to find an optimal solution than proving its optimality. When analyzing only instances that had the longest computing time, we observed that close to $100\%$ of computing time is used to obtain an optimal solution that satisfies one or two more customers than the incumbent solution. 

\begin{figure}[h!]
    \centering
    \begin{subfigure}[b]{0.46\textwidth}
    \includegraphics[width=0.92\linewidth]{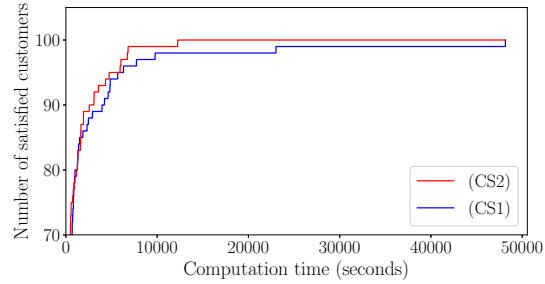}
    \end{subfigure}
    \caption{A graphical illustration of the solutions obtained by the MIP models for the group \texttt{st}, over the set of instances with $1000$ customers. The plot reports the number of solved instances within a time limit.}
    \label{fig:st-n1000}
\end{figure}

Also, as can be seen in Fig.~\ref{fig:st-n1000}, there were instances where (CS1) took close to~13h30m to finish where (CS2) can finish any instance in about 3h30m. Indeed, in a particular subset of a few instances, (CS1) cannot find an optimal solution in a reasonable amount of time.

All the results previously presented include the proposed preprocessing procedure of Section~\ref{sec:preprocessing} and the Gurobi's presolve engine (\ie, a collection of preprocessing techniques that reduce the size of the given model and also improve the strength of the model). 

\begin{figure}[ht!]
    \centering
    \begin{subfigure}[b]{0.48\textwidth}
    \centering
    \includegraphics[width=0.92\linewidth]{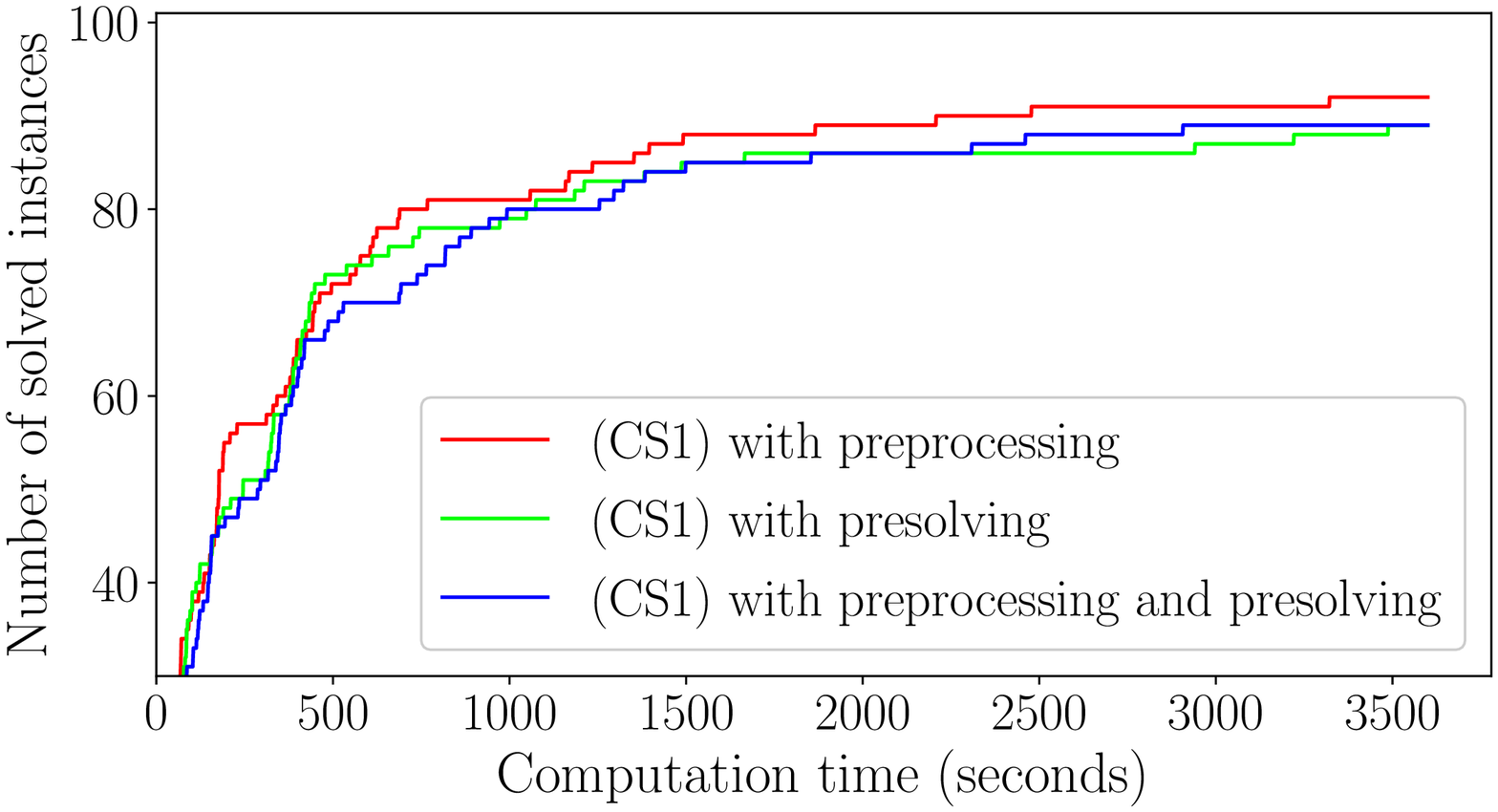}
    \end{subfigure}
    \begin{subfigure}[b]{0.48\textwidth}
    \centering
    \includegraphics[width=0.92\linewidth]{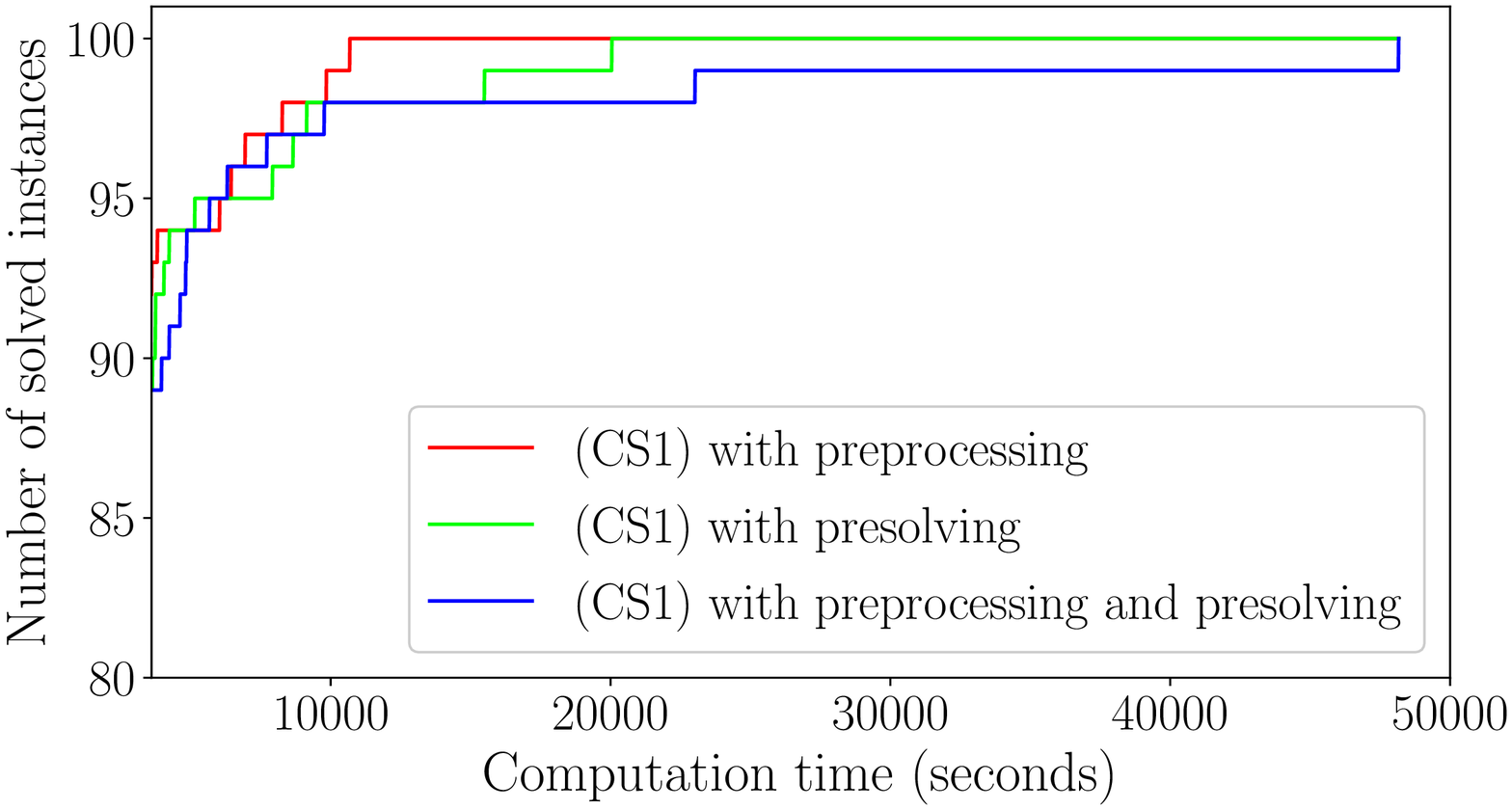}
    \end{subfigure}
    \begin{subfigure}[b]{\textwidth}
    \end{subfigure}
    \begin{subfigure}[b]{0.48\textwidth}
    \centering
    \includegraphics[width=0.92\linewidth]{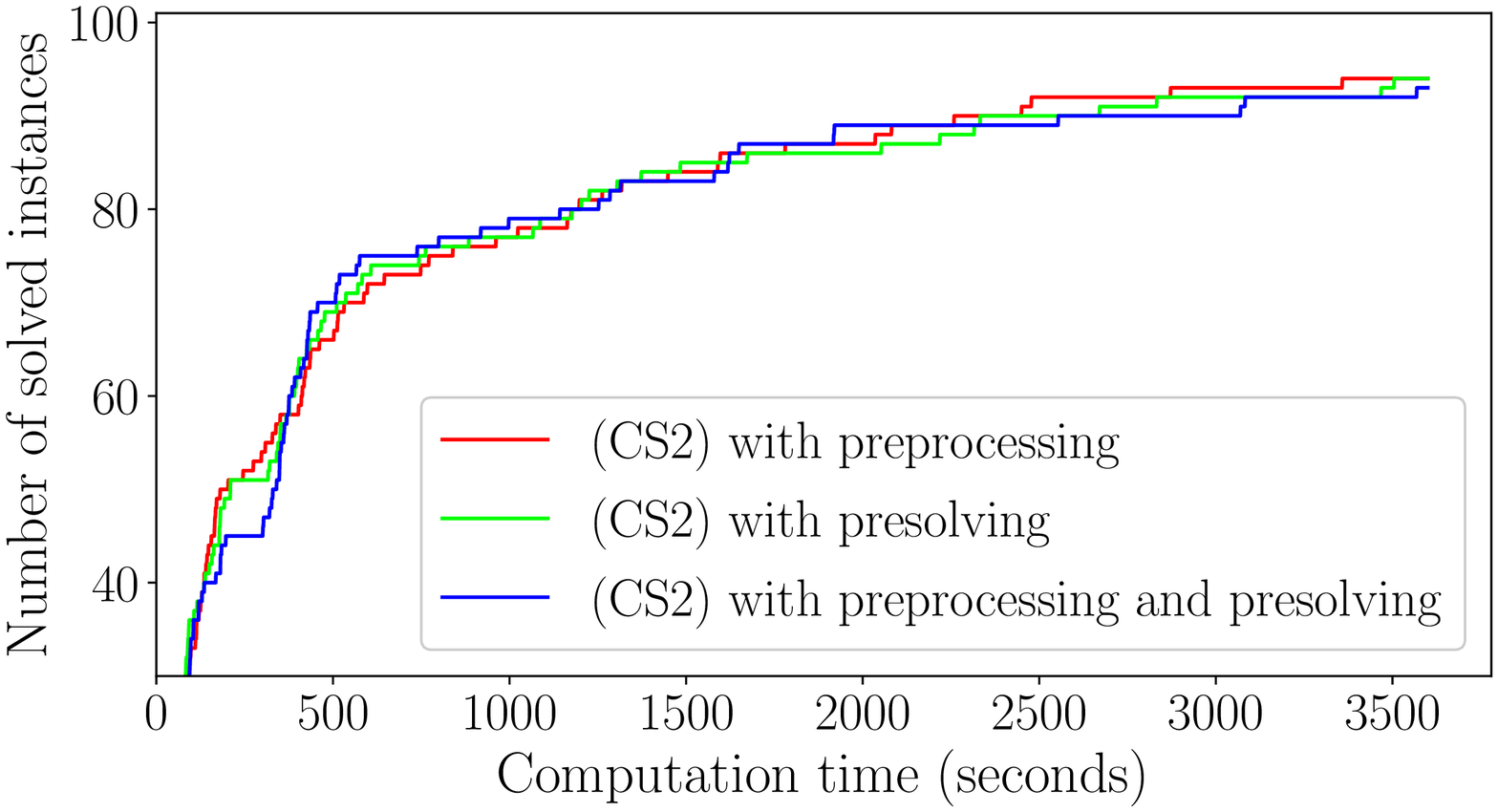}
    \end{subfigure}
    \begin{subfigure}[b]{0.48\textwidth}
    \centering
    \includegraphics[width=0.92\linewidth]{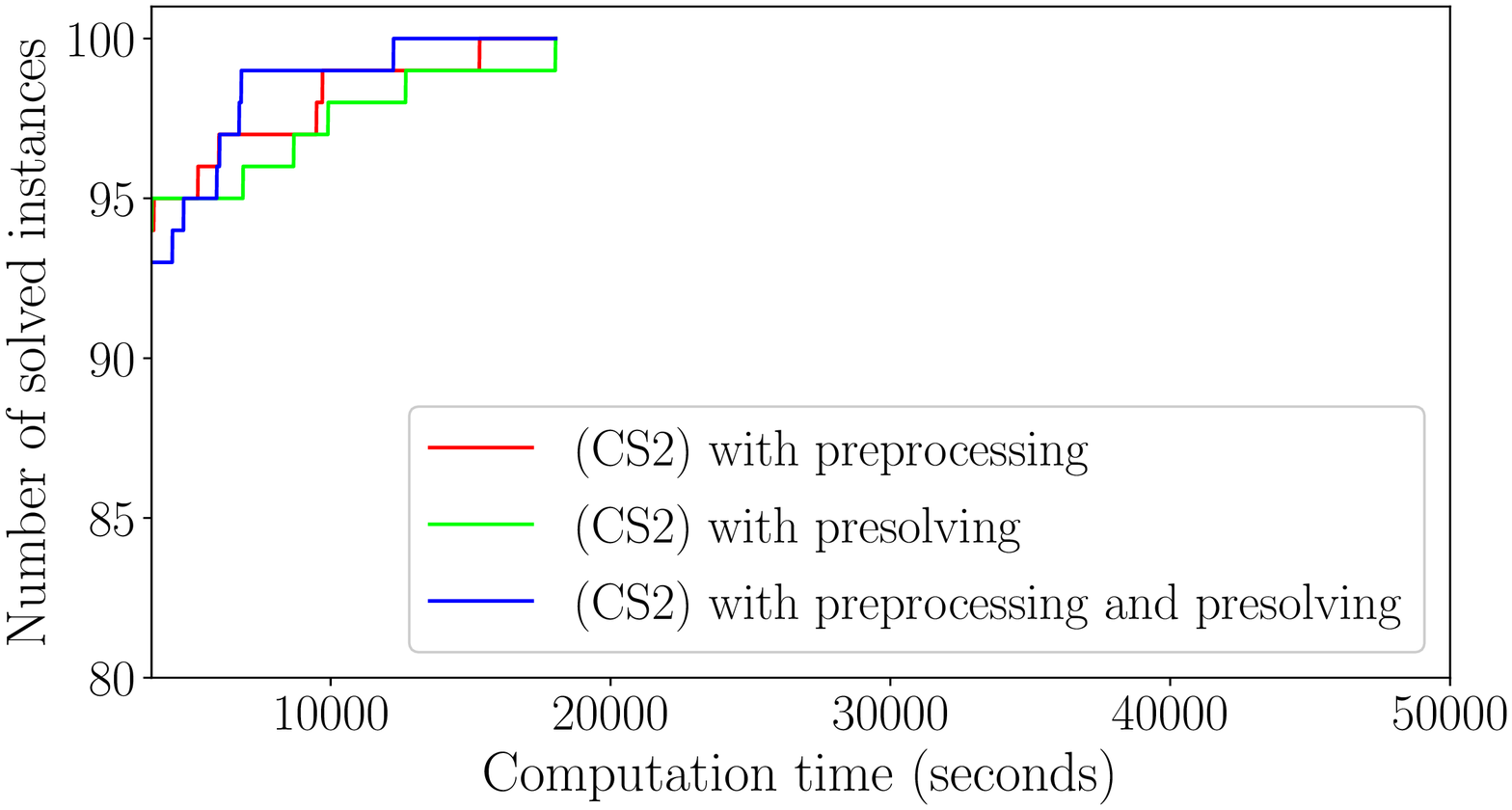}
    \end{subfigure}
    \caption{Performance of combinations between our preprocessing procedure and the Gurobi's presolve engine. In the legend labels, preprocessing means that the instance size has been reduced by using the proposed preprocessing procedure, while presolving means that the presolve phase of the Gurobi was enabled during the computational experiments.}
    \label{fig:preprocessing}
\end{figure}

In order to evaluate the impact of the proposed preprocessing procedure, Fig.~\ref{fig:preprocessing} illustrates the performance on three possible configurations. The results of (CS1) are intriguing, given that the configuration with only the preprocessing procedure performed better than the other two configurations. Since Gurobi is a commercial solver, it is difficult to understand this behavior. On the other hand, the results of (CS2) show that the configurations performed similarly. 

Checking the number of rows and columns of the models obtained, we observed that the reduction achieved by preprocessing the instance always kept $4$ rows and $6$ columns more than the presolve engine. We also observed that the number of rows and columns were reduced to less than $46\%$ and $65\%$ from the original values, respectively. As can be seen in the plots in Fig.~\ref{fig:preprocessing}, the preprocessing procedure enable relevant reductions of the model sizes. And, even though it presented a similar performance than the presolve engine, it is also useful for heuristic design.

\subsection{Matheuristic experiments}

Table~\ref{tab:matheuristics-avg-results} presents a comparison of the average values for solutions obtained by the matheuristics, where every execution was performed during $10$ minutes. In this table, column \emph{\small \#Opt} also reports, per instance set, the number of instances where we obtained an optimal solution. In addition, columns \emph{\small Avg. \#Impr}, \emph{\small Avg. satisfied customers}, \emph{\small Avg. gap~(\%)}, \emph{\small Avg. iterations} report the average number of times the search procedure improved the solution obtained, the average number of satisfied customers, the average relative gap (computed as \mbox{\small $100 \cdot (\UB-\LB)/\UB$}, where {\small $\LB$} is the lower bound and {\small $\UB$} is the best known upper bound computed by one of the MIP models), and the average of iterations performed by the matheuristic, respectively. Again, values after $\pm$ denote the standard deviation.
\begin{table}[ht!]
\centering
\caption{Aggregated matheuristic results for each instance set. In {\bf boldface}, the best number of instances where an optimal solution was found and the best average of relative gap value (when there is no tie).}
\label{tab:matheuristics-avg-results}
\resizebox{\textwidth}{!}{%
\begin{tabular}{cllrrrrrrrrrr}
\hline
\hline \vspace*{-0.3cm}\\
 &                     &  & \multicolumn{6}{l}{Greedy Randomized Adaptive Search Procedure (GRASP)} \\
\cline{4-9}  
 & Instance set        &  & \#Opt   & Avg. \#Impr      & Avg. construction   & Avg. satisfied customers & Avg. gap (\%)           & Avg. iterations \\
\hline
 & \texttt{st}-$n$1000 &  &      0  &  2.34 $\pm$ 0.89 & 391.44 $\pm$   4.77 &      398.78 $\pm$   4.19 &  \bf{1.353 $\pm$ 0.279} & 13.51 $\pm$ 1.30 \\
 & \texttt{st}-$n$2500 &  &      0  &  1.04 $\pm$ 0.20 & 468.41 $\pm$   5.91 &      478.27 $\pm$   4.67 &  \bf{2.497 $\pm$ 0.372} &  1.36 $\pm$ 0.48 \\
 & \texttt{st}-$n$5000 &  &      0  &  1.00 $\pm$ 0.00 & 525.25 $\pm$   5.81 &      535.50 $\pm$   4.37 &      3.195 $\pm$ 0.327  &  1.00 $\pm$ 0.00 \\
 & \texttt{ft}-$n$1000 &  & \bf{13} &  2.85 $\pm$ 0.93 & 462.76 $\pm$ 140.33 &      471.93 $\pm$ 135.94 &  \bf{1.726 $\pm$ 1.278} & 16.87 $\pm$ 9.77 \\
 & \texttt{ft}-$n$2500 &  &      0  &  1.19 $\pm$ 0.39 & 438.24 $\pm$ 176.27 &      487.40 $\pm$ 163.62 &  \bf{6.759 $\pm$ 1.454} &  1.51 $\pm$ 0.50 \\
 & \texttt{ft}-$n$5000 &  &      0  &  1.00 $\pm$ 0.00 & 387.39 $\pm$ 153.85 &      481.42 $\pm$ 158.89 &  \bf{9.045 $\pm$ 1.159} &  1.00 $\pm$ 0.00 \\
 & \texttt{fc}-$n$1000 &  &  \bf{5} &  3.35 $\pm$ 1.18 & 432.13 $\pm$ 122.14 &      442.10 $\pm$ 122.90 &  \bf{1.644 $\pm$ 0.989} & 18.44 $\pm$ 2.95 \\
 & \texttt{fc}-$n$2500 &  &      0  &  1.68 $\pm$ 0.55 & 429.48 $\pm$ 144.31 &      461.64 $\pm$ 145.63 &  \bf{6.410 $\pm$ 1.496} &  2.02 $\pm$ 0.51 \\
 & \texttt{fc}-$n$5000 &  &      0  &  1.00 $\pm$ 0.00 & 374.39 $\pm$ 145.85 &      449.97 $\pm$ 147.06 &     10.793 $\pm$ 1.597  &  1.00 $\pm$ 0.00\vspace*{-0.3cm} \\
\\
\hline
\hline \vspace*{-0.3cm}\\
 &                     &  & \multicolumn{6}{l}{Variable Neighborhood Search (VNS)} \\
\cline{4-9}
 & Instance set        &  & \#Opt  & Avg. \#Impr      & Avg. construction   & Avg. satisfied customers & Avg. gap (\%)           & Avg. iterations \\
\hline
 & \texttt{st}-$n$1000 &  &      0 &  1.57 $\pm$ 0.71 & 392.41 $\pm$   4.91 &      398.01 $\pm$   4.34 &       1.544 $\pm$ 0.368  & 23.27 $\pm$  0.96 \\
 & \texttt{st}-$n$2500 &  &      0 &  1.02 $\pm$ 0.14 & 469.90 $\pm$   5.90 &      478.07 $\pm$   4.71 &       2.538 $\pm$ 0.379  &  1.52 $\pm$  0.50 \\
 & \texttt{st}-$n$5000 &  &      0 &  1.00 $\pm$ 0.00 & 526.73 $\pm$   5.88 &      535.65 $\pm$   4.40 &   \bf{3.168 $\pm$ 0.361} &  1.00 $\pm$  0.00 \\
 & \texttt{ft}-$n$1000 &  &      3 &  1.80 $\pm$ 0.88 & 460.26 $\pm$ 139.17 &      469.43 $\pm$ 135.76 &       2.274 $\pm$ 1.356  & 29.74 $\pm$ 10.60 \\
 & \texttt{ft}-$n$2500 &  &      0 &  1.06 $\pm$ 0.24 & 438.36 $\pm$ 177.05 &      487.21 $\pm$ 163.79 &       6.817 $\pm$ 1.512  &  1.60 $\pm$  0.62 \\
 & \texttt{ft}-$n$5000 &  &      0 &  1.00 $\pm$ 0.00 & 388.12 $\pm$ 153.97 &      481.44 $\pm$ 159.78 &       9.097 $\pm$ 1.204  &  1.00 $\pm$  0.00 \\
 & \texttt{fc}-$n$1000 &  &      2 &  1.83 $\pm$ 1.05 & 431.82 $\pm$ 125.18 &      439.00 $\pm$ 123.13 &       2.390 $\pm$ 1.174  & 31.75 $\pm$  8.47 \\
 & \texttt{fc}-$n$2500 &  &      0 &  1.06 $\pm$ 0.28 & 433.91 $\pm$ 152.73 &      460.61 $\pm$ 147.11 &       6.742 $\pm$ 1.569  &  2.13 $\pm$  0.58 \\
 & \texttt{fc}-$n$5000 &  &      0 &  1.00 $\pm$ 0.00 & 388.84 $\pm$ 154.37 &      451.72 $\pm$ 148.13 &  \bf{10.473 $\pm$ 1.982} &  1.00 $\pm$  0.00\vspace*{-0.3cm} \\
\\
\hline
\hline \vspace*{-0.3cm}\\
 &                     &  & \multicolumn{6}{l}{Tabu Search (TS)} \\
\cline{4-9}
 & Instance set        &  & \#Opt  & Avg. \#Impr       & Avg. construction   & Avg. satisfied customers & Avg. gap (\%)           & Avg. iterations \\
\hline
 & \texttt{st}-$n$1000 &  &      0 &  3.66 $\pm$  1.45 & 392.41 $\pm$   4.91 &      396.77 $\pm$   4.44 &       1.850 $\pm$ 0.424 &  42.41 $\pm$ 14.93 \\
 & \texttt{st}-$n$2500 &  &      0 &  3.34 $\pm$  1.65 & 469.90 $\pm$   5.90 &      474.79 $\pm$   5.28 &       3.207 $\pm$ 0.595 &   8.71 $\pm$  4.34 \\
 & \texttt{st}-$n$5000 &  &      0 &  2.77 $\pm$  1.73 & 526.73 $\pm$   5.88 &      531.19 $\pm$   5.55 &       3.975 $\pm$ 0.591 &   6.79 $\pm$  3.58 \\
 & \texttt{ft}-$n$1000 &  &      3 &  6.59 $\pm$  3.11 & 460.26 $\pm$ 139.17 &      468.43 $\pm$ 136.27 &       2.528 $\pm$ 1.539 &  46.66 $\pm$ 15.60 \\
 & \texttt{ft}-$n$2500 &  &      0 & 21.89 $\pm$  9.70 & 438.36 $\pm$ 177.05 &      476.81 $\pm$ 163.64 &       8.998 $\pm$ 2.133 &  38.35 $\pm$ 18.50 \\
 & \texttt{ft}-$n$5000 &  &      0 & 43.72 $\pm$ 15.46 & 388.12 $\pm$ 153.97 &      464.89 $\pm$ 154.89 &      12.228 $\pm$ 1.700 &  75.33 $\pm$ 29.09 \\
 & \texttt{fc}-$n$1000 &  &      1 &  4.43 $\pm$  3.23 & 431.82 $\pm$ 125.18 &      438.21 $\pm$ 123.54 &       2.602 $\pm$ 1.283 & 104.07 $\pm$ 67.31 \\
 & \texttt{fc}-$n$2500 &  &      0 & 10.38 $\pm$ 10.23 & 433.91 $\pm$ 152.73 &      451.72 $\pm$ 146.27 &       8.650 $\pm$ 2.337 &  21.06 $\pm$ 20.14 \\
 & \texttt{fc}-$n$5000 &  &      0 & 22.63 $\pm$ 20.51 & 388.84 $\pm$ 154.37 &      427.93 $\pm$ 143.67 &      15.386 $\pm$ 5.206 &  41.46 $\pm$ 38.74\vspace*{-0.3cm} \\ \\
\hline
\hline
\multicolumn{9}{l}{\small \emph{Note.} Avg., average; \#Opt, number of optimal solutions; \#Impr, number of times the search process improved the solution.}
\end{tabular}%
}
\end{table}

From this table, it is possible to notice that the construction procedure obtained high-quality solutions for the instances with $1000$ customers. Besides that, the construction procedure was obtained good solutions in larger instances. Checking only the instances with~$5000$ customers, we observed that on average an initial solution is computed less than~$50$ seconds. Since only a single iteration was performed within 10 minutes, most of the computation time was spent in the local search. It is worth mentioning that the local search, in these cases, was interrupted due to the time limit. That is, our initial solution obtained good solution for large instances but the local search procedure is computationally expensive for such time limit in large instances. 

Comparing the matheuristics, GRASP performs better than the other two. Therefore, the proposed neighborhood structures to escape the highest peaks (local optimality) were not effective. This can be explained by the fact that it does not sufficiently decrease the number of satisfied customers in order to escape the highest peaks.

    \section{Conclusions}
\label{sec:conclusion}

In this paper, the {\probFullName} was solved by exact and heuristic approaches. First, we have proposed a preprocessing procedure to reduce the size of the instance. Next, we have described a mixed-integer programming (MIP) formulation based on network flow assignment, and we have proposed a new family of constraints to strengthen the formulation. Thus, we considered two models: (CS1), the original model; and (CS2), a strengthened version of~(CS1). We also have proposed three matheuristics, with an initial solution greedily built using linear programming, which are based on local search: a greedy randomized adaptive search procedure (GRASP); a variable neighborhood search (VNS); and a tabu search (TS). 
Computational experiments have been performed on the benchmark of instances to evaluate and compare the models and the matheuristics. 

In general, the MIP models perform better than the matheuristics. Except for two instance sets, the MIP models, on average, give less than $1\%$ of gap (results with computing-time limited to $10$ minutes). Both models, (CS1) and (CS2), have similar results. However, an interesting conclusion can be drawn from the computational time that is needed for obtaining an optimal solution. Model (CS1) has a much higher total computation time than~(CS2) in a particular set of instances. We observed that, in these cases, almost all the computing time needed is used to obtain an optimal solution that satisfies one or two customers more than the incumbent solution. We also observed that the proposed preprocessing procedure performs as well as the presolve engine of the commercial MIP solver used, with the additional advantage of being useful also for heuristic design.

The heuristic approaches, for the sets of instances with $5000$ customers, on average obtain a gap of less than $11\%$. Also, the matheuristics perform better than the MIP models for an instance set of $5000$ customers, whereas the MIP models on average obtain a gap greater than $39\%$ (and not finding a feasible solution for $35\%$ of the instances). Also, for sets of instances with $1000$ customers, the matheuristics on average obtains a gap less than $2\%$.

Finally, we believe another contribution of this paper is that many of the ideas used in the preprocessing procedure, the design of the MIP models, and the design of the heuristics can be used for a more general problem with more stations and customers with more demands.
    
    \section*{Acknowledgments}
    
    This study was financed in part by the Coordenação de Aperfeiçoamento de Pessoal de Nível Superior - Brasil (CAPES) - Finance Code 001. Supported by Grant {425340/2016-3}, 425806/2018-9, and 308689/2017-8, National Council for Scientific and Technological Development (CNPq). Supported by Grant 2015/11937-9, and 2017/23343-1, São Paulo Research Foundation (FAPESP).

    \bibliography{references.bib}
    \bibliographystyle{elsarticle-num}
    
\end{document}